\documentclass[journal]{IEEEtran}
\usepackage{cite}
\usepackage{amsmath,amssymb,amsfonts}
\usepackage{graphicx}
\usepackage{textcomp}
\usepackage[english]{babel}
\usepackage[normalem]{ulem}
\usepackage{amsmath}
\usepackage{graphicx,epstopdf}
\usepackage{hyperref}
\hypersetup{hidelinks}
\usepackage[english]{babel}
\usepackage[ruled,vlined]{algorithm2e}
\usepackage[table]{xcolor} 
\usepackage{amsmath}
\usepackage{booktabs}
\usepackage{hyperref}
\usepackage{graphicx}
\usepackage{amsmath,amssymb}
\usepackage{latexsym}
\usepackage{eurosym}
\usepackage{mathtools}
\usepackage{lscape} 
\usepackage{xcolor}
\usepackage{lipsum}
\usepackage{optidef}
\usepackage{multirow}
\usepackage{booktabs}
\usepackage{siunitx}

\usepackage{fontawesome}

\def\BibTeX{{\rm B\kern-.05em{\sc i\kern-.025em b}\kern-.08em
    T\kern-.1667em\lower.7ex\hbox{E}\kern-.125emX}}

\usepackage{amsthm}
\usepackage{graphicx}
\usepackage{subcaption}

\usepackage{orcidlink}
\newcommand{\tableref}[1]{\tablename{}~\ref{#1}}

\begin{document}

\title{Generalized Multi-Objective Reinforcement Learning with Envelope Updates in URLLC-enabled Vehicular Networks }
\author{Zijiang~Yan\orcidlink{0000-0002-7959-8329},~\IEEEmembership{Graduate Student Member,~IEEE} and
        Hina~Tabassum\orcidlink{0000-0002-7379-6949},~\IEEEmembership{Senior Member,~IEEE}%
\thanks{Z.Yan and H.Tabassum were with the Department
of Electrical Engineering and Computer Science, York University, Toronto,
ON, M3J 1P3 Canada e-mail: \{zjiyan,hinat\}@yorku.ca.}
\thanks{This research was supported by a Discovery Grant funded by the Natural Sciences and Engineering Research Council of Canada (NSERC). }
\thanks{A preliminary version of this work has been presented at the IEEE Global Communications Conference (GLOBECOM), 2022 \cite{10001396}}
}

\markboth{IEEE Transactions on Vehicular Technology, ~Vol.~1, No.~1, October~2025}%
{Shell \MakeLowercase{\textit{et al.}}: Bare Demo of IEEEtran.cls for IEEE Journals}

\maketitle
 
\begin{abstract}
We develop a novel multi-objective reinforcement learning (MORL) framework to jointly optimize wireless network selection and autonomous driving policies in a multi-band vehicular network operating on conventional sub-6GHz spectrum and Terahertz frequencies.  The proposed framework is designed to \textbf{(i)}~maximize the traffic flow and minimize collisions by controlling the vehicle's motion dynamics (i.e., speed and acceleration), and \textbf{(ii)}~enhance the ultra-reliable low-latency communication (URLLC) while minimizing handoffs (HOs). We cast this problem as a multi-objective Markov Decision Process (MOMDP) and develop solutions for both predefined and unknown preferences of the conflicting objectives. Specifically, we develop a novel envelope MORL solution which develops policies that address multiple objectives with unknown preferences  to the agent. While this approach reduces reliance on scalar rewards, policy effectiveness varying with different preferences is a challenge. To address this, we apply a generalized version of the Bellman equation and optimize the convex envelope of multi-objective Q values to learn a unified parametric representation capable of generating optimal policies across all possible preference configurations. Following an initial learning phase, our agent can execute optimal policies under any specified preference or infer preferences from minimal data samples.
 Numerical results validate the efficacy of the envelope-based MORL solution and demonstrate interesting insights related to the inter-dependency of vehicle motion dynamics, HOs, and the communication data rate. The proposed policies enable autonomous vehicles (AVs) to adopt
safe driving behaviors with improved connectivity.
\end{abstract}


\IEEEpeerreviewmaketitle

\raggedbottom
\section{Introduction}
\IEEEPARstart{F}{}acilitating  ultra-reliable and low-latency vehicle-to-infrastructure (V2I) communications is a fundamental prerequisite for the realization of autonomous and intelligent transportation systems. Different from throughput-oriented conventional communications, ensuring ultra-reliable low latency communications (URLLC) is  challenging as it relies on ensuring the signal-to-interference ratio (SINR), data rate, over-the-air/queuing latency, and decoding probability. Conventional radio frequency (RF) alone cannot efficiently meet the stringent URLLC requirement due to its limited coverage and narrow transmission bandwidths. In this context, 6G technology enables  combining the conventional sub-6GHz transmissions\footnote{We use sub-6GHz and RF communication interchangeably in this paper.} in conjunction with extremely high frequencies such as THz transmissions, where the former can compensate for the severe path-loss attenuation of THz transmission, and the latter can help overcome the RF spectrum congestion. 

{
On the other hand, the use of Deep Reinforcement Learning (DRL) is becoming critical for online decision making in highly random, mobile-oriented wireless environments.
In the context of V2I communications, a wealth of research has focused on improving network quality of service (QoS) (e.g., including transmission delay, link throughput, etc.) via DRL-based resource allocation \cite{ye2019deep,liang2019spectrum,xu2023deep}. This research has focused on considering subchannel and power allocation to improve V2I communication. 
In particular, the authors in \cite{ye2019deep,liang2019spectrum,yan2024hybrid} adopted deep-Q network (DQN) and multi-agent DQN to simultaneously improve the overall throughput of V2I links and the payload delivery rate of vehicle-to-vehicle (V2V) links.  Xu \textit{et al.} \cite{xu2023deep} derived the contribution-based dual-clip proximal policy to optimise V2I and V2V connections separately. However, their system model includes only a single BS where handovers (HOs) are not considered. 
}

Recently, the authors in \cite{hu2020dynamic,yu2015multi,devarajan2012energy, guo2020joint,khan2019reinforcement} formulated similar optimization tasks as multi-objective optimization problem (MOOP) and proposed to use multi-objective reinforcement learning (MORL) solutions. Hu \textit{et al.} \cite{hu2020dynamic} implemented Double-Loop Learning (DLL) to minimize the latency of real-time services transmission and maximize the throughput of non-instant services transmission.   In \cite{yu2015multi,devarajan2012energy}, the authors adopted the weighted Tchebycheff method and weighted-sum-MORL to maximize the fraction between the data rate and the power consumption, respectively. 
From \cite{guo2020joint}, Guo \textit{et al.} applied the Multi-Agent Proximal Policy Optimization (MAPPO) algorithm to address the joint handover and power allocation problem. In \cite{khan2019reinforcement}, Khan \textit{et al.} utilized the Asynchronous Advantage Actor-Critic (A3C) algorithm to devise a vehicle-RSU association policy, aiming to enhance the mobile user experience by maximizing   sum data rate of multiple AVs while ensuring a minimum level of service rate for all AVs.

Along another note, most of the existing research works in the transportation are focused on collision-avoidance \cite{wu2020driving,yan2023multi}, safe driving \cite{liu2020enhancing,yan2024hybrid}, and efficient fuel consumption \cite{liu2020enhancing, alizadeh2019automated, he2023towards}. For instance, in \cite{wu2020driving}, the authors applied a RL framework for faster travel and the reward is proportional to the AV's velocity along with a penalty for vehicle collision.  The action space included acceleration, deceleration, lane changes (LC), and maintain speed, whereas the state space was based on AVs' locations and their respective velocities. In \cite{liu2020enhancing}, the authors applied DDQN to enhance AVs' driving safety and fuel consumption. The state space included AVs' locations, fuel consumption, and velocities, whereas the actions included speeds and LC of AVs.  In \cite{alizadeh2019automated}, the authors applied the Intelligent Driver Model (IDM) and Minimizing Overall Braking Induced by Lane Change (MOBIL) to control steering and lane change. The proposed reward design encourages long traffic, high speed and discourages unnecessary LC. Authors in \cite{he2023towards} introduced multi-objective actor-critic to improve the tradeoff between energy consumption and travel efficiency of AVs. The authors derived MO actor-critic to optimize two objectives.
{
Recently, a few optimization-based research works have started looking into this problem where the  AV velocity has been optimized using simplified car-following models \cite{Shoaib2023Macroscopic,Shoaib2023Optimization,Saeidi2023Tractable,pei2024detection,pei2025deep}, but these models are computationally expensive, do not learn from past experiences thus not relevant for faster decision-making, and are not applicable to dynamic traffic scenarios with various control and communication parameters.
}
\begin{table*}
\centering
\resizebox{\linewidth}{!}{
\begin{tabular}{|c|c|c|c|c|c|c|c|c|c|c|}
\hline
\multirow{2}{*}{\textbf{Ref.}} & 
\multicolumn{4}{|c|}{\textbf{Autonomous Driving}} & 
\multicolumn{4}{|c|}{\textbf{Vehicular Communication}} & 
\multirow{2}{*}{\begin{tabular}[c]{@{}c@{}}ML \\ Optimization \\ Method\end{tabular} } & 
\multirow{2}{*}{\begin{tabular}[c]{@{}c@{}}Computational\\ Complexity\end{tabular}} \\
\cline{2-9}
 & \begin{tabular}[c]{@{}c@{}}Collision\\ Avoidance\end{tabular} & 
 \begin{tabular}[c]{@{}c@{}}Speed\\ Management\end{tabular} & 
 \begin{tabular}[c]{@{}c@{}}Driving\\ Behaviour\end{tabular} & 
 \begin{tabular}[c]{@{}c@{}}Lane\\ Changes\end{tabular} & 
 \begin{tabular}[c]{@{}c@{}}Mobility Aware\\ URLLC\end{tabular} & 
 \begin{tabular}[c]{@{}c@{}}Handover\\ Management\end{tabular} & 
 \begin{tabular}[c]{@{}c@{}}Interference\\ Aware\end{tabular} & 
 \begin{tabular}[c]{@{}c@{}}Network\\ Availability\end{tabular} & &  \\
\hline
\multirow{1}{*}{\cite{10001396}} & $\checkmark$ & $\checkmark$ & $\checkmark$ & {} & {} & $\checkmark$ & $\checkmark$ & \begin{tabular}[c]{@{}c@{}} RF  + THz\\ Multibands \end{tabular} & Q + DQN  & Low \\
\hline
\multirow{1}{*}{\cite{liu2020enhancing}} & $\checkmark$ & $\checkmark$ & {} & $\checkmark$ & {} & {} & {} & AGWN & DQN + DDQN & Low \\
\hline
\multirow{1}{*}{\cite{ye2019deep}} & {} & {} & {} & {} & {} & {} & $\checkmark$ & \begin{tabular}[c]{@{}c@{}} Single Cell \\ system 2 GHz \end{tabular} & DQN & Low \\
\hline
\multirow{1}{*}{\cite{xu2023deep}} & {} & {} & {} & {} & $\checkmark$ & {} & $\checkmark$ & \begin{tabular}[c]{@{}c@{}} Multi-Platoon \\ Vehicular \end{tabular}   & MORL: CD-PPO & High \\
\hline
\multirow{1}{*}{\cite{he2023towards}} & $\checkmark$ & $\checkmark$ & {} & $\checkmark$ & {} & {} & {} & RF & MORL: MOAC & High \\
\hline
\multirow{1}{*}{\cite{song2022evolutionary}} & {} & {} & {} & {} & {} & $\checkmark$ & $\checkmark$ & \begin{tabular}[c]{@{}c@{}} UAV Assisted \\ MEC \end{tabular} & EMORL & High \\
\hline
\multirow{1}{*}{\cite{li2018urban}} & $\checkmark$ & $\checkmark$ & {} & $\checkmark$ & {}  & {}  & {}  & AGWN &  \begin{tabular}[c]{@{}c@{}} MORL: thresholded \\ lexicographic \end{tabular}   & High \\
\hline
This paper & $\checkmark$ & $\checkmark$ & $\checkmark$ & $\checkmark$ & $\checkmark$ & $\checkmark$ & $\checkmark$ & \begin{tabular}[c]{@{}c@{}} RF  + THz\\ Multibands \end{tabular} & \begin{tabular}[c]{@{}c@{}} MORL\\MO-DDQN-Envelope \end{tabular} & Low \\
\hline
\end{tabular}
}
\caption{Comparison between Related Works and This Work}
\label{tab:comparison_benchmarks}
\end{table*}
To date, none of the existing research works \cite{ye2019deep,xu2023deep,liang2019spectrum, guo2020joint,khan2019reinforcement,hu2020dynamic,wei2021multi,yu2015multi,devarajan2012energy,wu2020driving, liu2020enhancing,alizadeh2019automated,he2023towards,Shoaib2023Macroscopic,Shoaib2023Optimization,Saeidi2023Tractable} have considered the inter-dependency of the AV motion dynamics to wireless data rates.  

Different from the existing research, {our contributions can be explained from two perspectives, i.e., \textbf{(1)} jointly optimizing both autonomous driving policies and telecommunication rewards \textit{with and without predefined preferences}, and \textbf{(2)} proposing a novel multi-objective \textit{MO-DDQN-Envelope} reinforcement learning solution that enables optimizing policies across varying preferences (i.e., without predefined preferences) in the multi-objective domain.}  { \tableref{tab:comparison_benchmarks}  summarizes the distinction of existing research in terms of computational complexity, methodology, and performance metrics.} Our contributions can be summarized as follows:
\begin{itemize}
    \item We develop an MORL framework to design joint network selection and autonomous driving policies in a multi-band vehicular network (VNet). The objectives are to \textbf{(i)} maximize the traffic flow and minimize collisions by controlling the vehicle's motion dynamics (i.e., speed and acceleration) from a transportation perspective, and \textbf{(ii)} maximize the data rates and minimize handoffs (HOs) by jointly controlling the vehicle's motion dynamics and network selection from telecommunication perspective. We consider a novel reward function that maximizes data rate and traffic flow, ensures traffic load balancing across the network,  penalizes HOs, and unsafe driving behaviors. 
    \item The considered problem is formulated as a multi-objective Markov decision process (MOMDP) that has two-dimensional action space and rewards consist of  telecommunication and autonomous driving utilities. 
    We jointly optimizing both autonomous driving policies and telecommunication rewards \textit{with and without predefined preferences} using multi-objective reinforcement learning (MORL) approaches.

    \item We propose a novel multi-objective \textit{MO-DDQN-Envelope} reinforcement learning solution that enables optimizing policies across varying preferences (i.e., without predefined preferences) in the multi-objective domain. Our framework simultaneously optimizes telecommunication and transportation objectives in dynamic environments, balancing trade-offs like collision avoidance, velocity management, handover optimization, and network availability.
Unlike scalarized methods, our \textbf{MO-DDQN-Envelope} approach dynamically adjusts preferences, mitigating biases and errors introduced by fixed weightings.

\item {We develop  a novel simulation testbed that emulates multi-band wireless network-enabled VNet \textit{RF-THz-Highway-Env} based on \textit{highway-env} \cite{highway-env}. This test environment  not only inherits the advantages of autonomous driving, and lane changes on the highway from  \cite{highway-env}, but also implements  RF/THz  channel propagation modeling, network selection, and HO control.} 

\item Numerical results shows that the proposed solution outperforms weighted sum-based MORL solutions with DQN by $12.7\%$, $18.9\%$, and $12.3\%$ on average transportation reward, average communication reward, and average HO rate, respectively.
\end{itemize}

The rest of this work is organized as follows. Section \ref{systemmodel} shows the system model, and Section~\ref{Background} provides MOMDP formulation. Section~\ref{morl} introduces the proposed solution. The simulations are presented in Section \ref{simu}, and Section \ref{conclude} concludes this research work.

\section{System Model and Assumptions}
\label{systemmodel}
\begin{figure}
\includegraphics[width=1\linewidth]{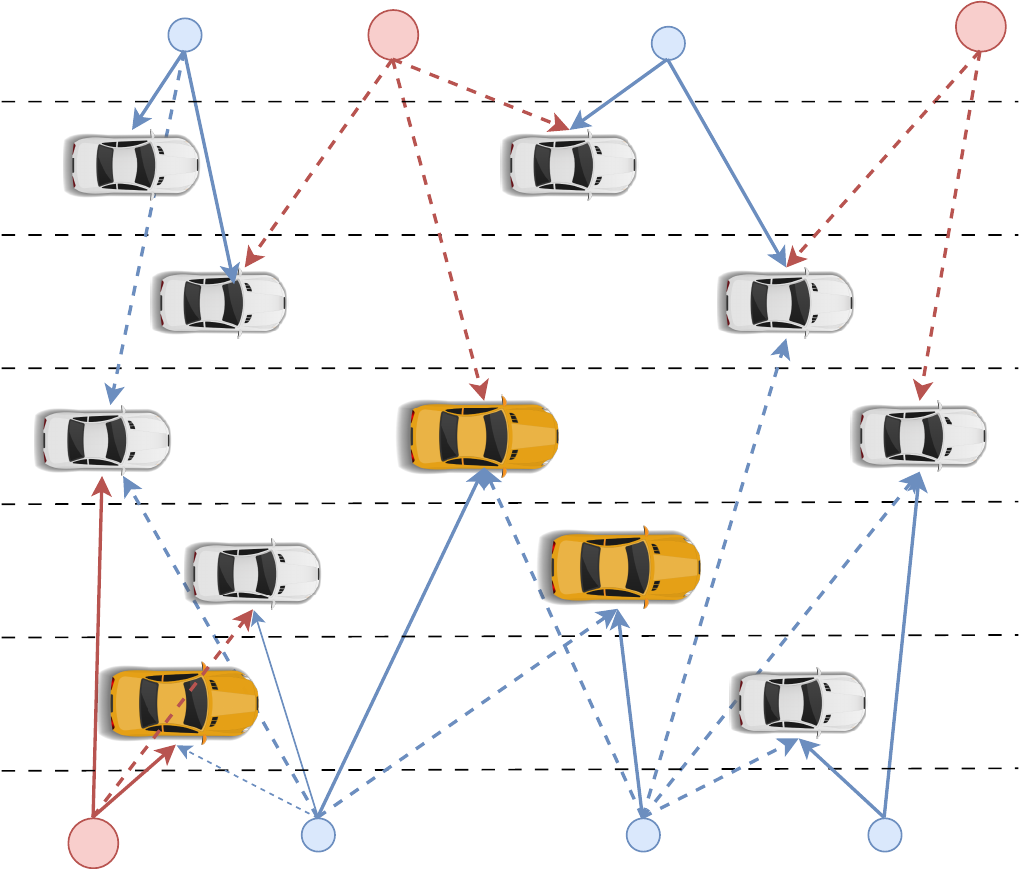}
\caption{The  diagram illustrates the structure of the multi-band VNet model. The blue and red circles represent TBSs and RBSs, respectively. The solid and dashed lines represent desired signal links and interference links, respectively.}
\label{fig:BS_distributions}
\end{figure}

{A multi-band downlink network comprising {  $n_R$ Radio Frequency Base Stations (RBSs) and $n_T$ Terahertz Frequency Base Stations (TBSs)} is considered. A multi-vehicle network is also considered, with a multi-lane road comprising $N_L$ lanes. $M$ AVs receive information from the BSs (deployed alongside the road) through V2I communications (as depicted in Figure \ref{fig:BS_distributions}). Each AV is permitted to associate with only one BS at a time, regardless of whether the BS is an RBS or TBS. The on-board units (OBUs) on the AVs receive real-time information from the VNet, including the velocity, acceleration, and lane position of surrounding vehicles. Each RBS and TBS has a bandwidth available to it, represented by $W_R$ and $W_T$, respectively. Each RBS and TBS is capable of supporting a maximum number of AVs, denoted by $Q_R$ and $Q_T$, respectively. All AVs are equipped with a single antenna.}

{Compared to the Gipps' model~\cite{Gipps1981behavioural} and learning-based approaches~\cite{Zhang2022generative}, we adopt a transportation model that combines \textbf{Kinematics + IDM + MOBIL}, which is widely used in autonomous driving decision-making, as in~\cite{alizadeh2019automated}. This hybrid model provides improved computational efficiency and adaptability to MORL, outperforming the Gipps and learning-based models in these aspects.
For the telecommunication model, in contrast to RF-only 5G V2I, THz-only 6G V2I~\cite{10001396}, and RIS-assisted models~\cite{Wei2023}, we adopt a hybrid multiband RF–THz V2I framework~\cite{hossan2021mobility}. This model offers superior performance in terms of throughput, latency, reliability, mobility management, and HOs control.
}

\subsection{Downlink V2I Data Transmission  Model}
\label{tele_model}
The signal transmitted by the RBS is subject to path-loss and short-term channel fading. Subsequently, the signal-to-interference-plus-noise ratio (SINR) of the $j$-th AV from $i$-th RBS is given as \cite{saeidi2024molecular, hossan2021mobility}:
\begin{equation}{\label{sinr_rf}}
    \mathrm{SINR}^{\mathrm{RF}}_{ij} = \frac{P_{R}^{\mathrm{tx}}\:G_{R}^{\mathrm{tx}}\:G_{R}^{\mathrm{rx}} \left(\frac{c}{4\pi f_{R}} \right)^2 H_{i} }{r_{ij}^{\alpha}\left(\sigma^2 + I_{R_j}\right)},
\end{equation}
{where \(P_{R}^{\text{tx}}\), \(G_{R}^{\text{tx}}\), \(G_{R}^{\text{rx}}\), \(c\), \(f_{R}\), \(r_{ij}\), and \(\alpha\) represent the transmit power of the RBSs, the gain of the transmitting antenna, the gain of the receiving antenna, the speed of light, the RF carrier frequency in GHz, the distance between the \(j\)-th AV and the \(i\)-th RBS, and the path-loss exponent, respectively.
{ Note that $r_{ij} = {(d_{ij}^2+h_{ij}^2)}^{1/2},$ where $d_{ij}$ is the 2D distance between the $j$-th AV and $i$-th BS and $h_{ij}$ is the transmit antenna height.} 
Furthermore, \(H_i\) denotes the exponentially distributed channel fading power observed by the \(j\)-th AV from the \(i\)-th RBS, \(\sigma^2\) is the power of the thermal noise at the receiver, and \(I_{R_j}\) is the cumulative interference experienced by the \(j\)-th AV from other interfering RBSs.
}
$
I_{R_j} = \sum_{k\neq i} P_{R}^{\mathrm{tx}} \gamma_{R} r_{kj}^{-\alpha} H_{k}
$
where $r_{kj}$ is the distance between the $k$-th interfering RBS and the $j$-th AV, $H_{k}$ is the power of fading from the $k$-th interfering RBS to the $j$-th AV, and $\gamma_R = G_{R}^{\mathrm{tx}} \: G_{R}^{\mathrm{rx}} \left( {c}/{4\pi f_{R}} \right)^2$. 

{
In the context of a Terahertz (THz) network, where molecular absorption significantly impacts signal propagation, the significance of line-of-sight (LOS) transmissions over non-line-of-sight (NLOS) transmissions is dominant. Consequently, the SINR for a given $j$-th AV can be modeled as follows:
}
\begin{equation}
        \mathrm{SINR}^{\mathrm{THz}}_{ij} = \frac{G_{T}^{\mathrm{tx}} G_{T}^{\mathrm{rx}} \left(\frac{c}{4\pi f_{T}} \right)^2 P_{T}^{\mathrm{tx}}\: \mathrm{exp}(-K_a(f_T) r_{ij})r_{ij}^{-2} }{N_{T_j} + I_{T_j}},
\end{equation}
{where} $ 
G_{T}^{\text{tx}}, G_{T}^{\text{rx}}, P_{T}^{\text{tx}}, f_{T}, r_{ij}, \text{ and } K_{a}(f_T) $
represent the transmit antenna gain of the TBS, the receiving antenna gain of the TBS, the transmit power of the TBS, the THz carrier frequency, the distance between the \(j\)-th AV and the \(i\)-th TBS, and the molecular absorption coefficient of the transmission medium, respectively \footnote{For the sake of brevity, the argument of \( K_{a}(f_T) \) will henceforth be omitted in this study.}.  It is important to note that:
$
G_{T}^{\text{rx}}(\theta) \text{ and } G_{T}^{\text{tx}}(\theta)
$
denote the antenna gains at the receiver and transmitter sides corresponding to the boresight direction angle \( \theta \in [-\pi, \pi) \). The beamforming gain from the main and side lobes of the TBS transmitting antenna is subsequently defined as,
\begin{equation}
\label{eq:gain2}
  G_{T}^{q}\left(\theta\right) =
    \begin{cases}
      G^q_{\mathrm{max}} & \mid \theta \mid \leq w_{q}\\ 
      G^q_{\mathrm{min}} & \mid \theta \mid > w_{q}
    \end{cases},  
\end{equation}
where the superscript $q$ is used to indicate the transmit/ receive antenna, i.e., $q \in \{\mathrm{tx,rx}\}$, $w_{q}$ is the beamwidth of the main lobe, $G^q_{\mathrm{max}}$ and $G^q_{\mathrm{min}}$ are the beamforming gains of the main and side lobes, respectively. {We assume that AVs can align the receiving beam with the TBS transmit beam using beam alignment techniques.} For the alignment between the user and interfering TBSs, we define a random variable {$\Theta$},  $\Theta \in \{G^{\mathrm{tx}}_{\mathrm{max}}G^{\mathrm{rx}}_{\mathrm{max}},G^{\mathrm{tx}}_{\mathrm{max}}G^{\mathrm{rx}}_{\mathrm{min}},G^{\mathrm{tx}}_{\mathrm{min}}G^{\mathrm{rx}}_{\mathrm{max}},G^{\mathrm{tx}}_{\mathrm{min}}G^{\mathrm{rx}}_{\mathrm{min}}\},$ and the respective probability for each case is $F_{\mathrm{tx}}F_{\mathrm{rx}}$, $F_{\mathrm{tx}}(1-F_{\mathrm{rx}})$, $(1-F_{\mathrm{tx}})F_{\mathrm{rx}}$, and $(1-F_{\mathrm{tx}})(1-F_{\mathrm{rx}})$, where $F_{\mathrm{tx}} = \frac{\theta_{\mathrm{tx}}}{2\pi}$ $F_{\mathrm{rx}} = \frac{\theta_{\mathrm{rx}}}{2\pi}$ { with $\theta_{\mathrm{tx}}, \theta_{\mathrm{rx}}$ being the beamwidth on the transmitter and receiver antenna, respectively}. Without loss of generality, we consider negligible side lobe gains. Thus, the cumulative interference $I_{T}$ between AV and the interfering TBS is given as
$
 I_{T} = \sum_{k\neq i}\gamma_T \:P_{T}^{\mathrm{tx}}\:F_{\mathrm{tx}}F_{\mathrm{rx}} 
r_{kj}^{-2} \mathrm{exp}(-K_a \:{r_{kj}}),
$
where $\gamma_T=G_{T}^{\mathrm{tx}} G_{T}^{\mathrm{rx}} \left(\frac{c}{4\pi f_{T}} \right)^2$. The cumulative thermal and molecular absorption noise is thus given as:
\begin{equation}
\begin{multlined}
    N_{T_j}= N_{0} +  \:P_{T}^{\mathrm{tx}} \gamma_T \:{r_{ij}^{-2}} \:(1-e^{-K_a \:{r_{ij}}})+ \\
\sum_{k \neq i} \gamma_T F_{\mathrm{tx}}F_{\mathrm{rx}} \:P_{T}^{\mathrm{tx}} \:{r_{kj}^{-2}}(1-e^{-K_a \:{r_{kj}}}).
\end{multlined}
\end{equation}

The traditional data rate relies on Shannon's capacity, which can be attained as the block-length of channel codes approaches infinity. Nevertheless, to prevent prolonged transmission delays in URLLC, the block length must be limited. Consequently, Shannon's capacity cannot be realized due to the presence of a non-zero decoding error probability~\cite{she2021tutorial}.  From \cite{9170614}, the achievable rate in the short block-length regime over an AWGN channel can be approximated as:
\begin{equation}
\label{eq:achievable_rate}
 R_{ij} = \frac{W_j}{\ln{2}} \left[ \ln(1+\mathrm{SINR}_{ij}) -\sqrt{\frac{V}{L_B}}f_Q^{-1}(\epsilon_c) \right]   
\end{equation}
where $W_j,L_B,\epsilon_c,f_Q^{-1}(\cdot),V$ are the transmission bandwidth of BS $j$, blocklength, decoding error probability, {the inverse $Q$ function, and the} channel dispersion, respectively. $V$ can be calculated by $1 - \frac{1}{(1+ \mathrm{SINR}_{ij})^2}$. 
{Given that $D_t$ time to transmit $L_B$ symbols, the time  and frequency resources can be computed by $ D_t W = L_B$. where $W=W_R$ for RBSs and $W= W_T$ for TBSs.} As the block length $L_B$ approaches infinity, the achieve rate in (\ref{eq:achievable_rate}) reaches Shannon's capacity. 

Each AV maintains a list of the BSs in terms of the achievable data rate $R_{ij}$ and then informs these BSs.  Consequently, each BS can calculate the possible AV associations at each time instance denoted by $n_i$. Then, the AV collects the traffic load information from these BSs (i.e.,  the number of AVs associated with each BS $n_i$). 
Based on the quota of each BS $i$, $Q_i \in [Q_R, Q_T]$, each AV computes a \textit{weighted data rate metric} that encourages traffic load balancing at each BS and discourages unnecessary HOs, i.e.,
\begin{equation}
\label{eq:weight_data_rate}
    \text{WR}_{ij} = \frac{R_{ij}}{\min \left(Q_i, n_i \right)} (1 - \mu)
\end{equation}
where $\mu$ denotes the HO penalty to discourage unnecessary HOs {that is defined as follows:} 
\begin{equation}
{
\mu = 
\begin{dcases}
0.1 ,& \text{if switch to a RBS} \\
0.5, & \text{if switch to a TBS } \\
0, & \text{keep previous BS} 
\end{dcases}
}   
\end{equation}
 As AVs traverse the corridor, they transition from one BS to another, which is referred to as a HOs. We distinguish between two types of HOs: horizontal and vertical. Horizontal HO denotes the AV connection shifting from one BS of the same type to another. In contrast, vertical HO pertains to the scenario where the AV connection transitions from one specific type of BS to a distinct type of BS, such as moving from an RBS to a TBS. It is evident that frequent HOs can have a significant impact on the data rate that AVs receive, due to the inherent latency and failure rates associated with HOs. In this paper, we propose the introduction of a penalty, denoted by the parameter $\mu$, which is designed to discourage HOs. This penalty is higher for TBSs and lower for RBSs, reflecting the fact that THz transmission is limited to a relatively short distance, rendering it more vulnerable to unnecessary HOs. 

Then, each AV prepares a sorted list of BSs offering the best weighted data rates $\text{WR}_{ij}$ and associates to those that can fulfill the data rate requirement of the AV given by $R_{\mathrm{th}}$. 

\subsection{Transportation Model}
\label{transportation_model}
We categorize $M$ AVs into two groups: target vehicles, denoted as $M_1$, and surrounding vehicles, denoted as $M_2$. 

{ Following \cite{polack2017kinematic, highway-env}, we update the real-time physical location of all AVs using the \textit{Kinematics} model \cite{polack2017kinematic}. Assuming only the front wheels can be steered, the dynamics of each AV \(j\), \(j \in M\), are described as:

\begin{equation}
\label{eq:kinematics}
    \frac{\partial x_j}{\partial t} = v_j \cos (\psi_j + \beta_j), \quad
    \frac{\partial y_j}{\partial t} = v_j \sin (\psi_j + \beta_j), \quad
    \frac{\partial v_j}{\partial t} = a_j,
\end{equation}

where \((x_j, y_j)\) represents the position of AV \(j\), \(v_j\) its velocity, \(a_j\) its commanded acceleration, \(\psi_j\) its heading angle, and \(\beta_j\) the slip angle at the vehicle's center of gravity. The slip angle \(\beta_j\) is given by:
\begin{equation}
    \beta_j = \arctan \left( \frac{\tan \delta_j^\mathrm{fa}}{2} \right),
\end{equation}
where \(\delta_j^\mathrm{fa}\) is the steering angle of the front wheels.

The heading dynamics of AV \(j\) depend on its control model:
\begin{equation}
\frac{\partial \psi_j}{\partial t} = 
\begin{cases}
    K_j^{\psi} \left[ \psi_{L_j} + \arcsin \left( \frac{\tilde{v}_{j,y}}{v_j} \right) - \psi_j \right], & \text{if } j \in M_1, \\
    \frac{v_j}{l_j} \sin \beta_j, & \text{if } j \in M_2,
\end{cases}
\end{equation}
where \(l_j\) is the half-length of AV \(j\), \(\psi_{L_j}\) is the desired heading, and \(K_j^{\psi}\) and \(K_j^{y}\) are control gains. The term \(\tilde{v}_{j,y}\), representing lateral velocity adjustment, is given as:
\[
\tilde{v}_{j,y} = K_j^{y} (y_{L_j} - y_j),
\]
with \(y_{L_j}\) being the lateral position of the desired lane.

The acceleration \(a_j\) for AV \(j\) is defined as:
\begin{equation}
    a_j = 
\begin{cases}
    K_0^v (v_r - v_j), & \text{if } j \in M_1, \\
    a_c - a_c \left[ \left(\frac{|v_j|}{v_0} \right)^{\delta_a} + \left(\frac{\hat{d}_j}{d_j}\right)^2 \right], & \text{if } j \in M_2,
\end{cases}
\end{equation}
where \(v_r\) is the desired speed, \(K_0^v\) is the speed control gain, \(a_c\) is the maximum acceleration, \(v_0\) is the desired velocity, \(\delta_a\) is the acceleration reduction factor, and \(d_j\) is the distance to the front AV. The desired gap \(\hat{d}_j\) is computed as:
\begin{equation}
    \hat{d}_j = d_0 + \max \left( 0, T v_j + \frac{v_j \Delta v_j}{2 \sqrt{a_c b_c}} \right),
\end{equation}
where \(d_0\) is the minimum distance in stopped traffic, \(T\) is the safe time gap, \(\Delta v_j\) is the relative velocity to the front vehicle, and \(b_c\) is the comfortable braking deceleration.

For AVs in \(M_2\), discrete lane changes are determined by the MOBIL model \cite{kesting2007general, leurent2020safe}, where an AV decides to change lanes when (1) AVs are safe to cut-in another lane as $\tilde{a}_j \geq -b_\text{safe}$. (2) there is an incentive benefit if the target AVs and followers satisfies $\tilde{a}_j - a_j + p \left( \tilde{a}_o - a_o + \tilde{a}_n - a_n \right) \geq \Delta a_\text{th}$.
where \(-b_\text{safe}\) is the maximum braking imposed on the new following AV, \(p\) is the politeness coefficient, \(\Delta a_\text{th}\) is the minimum acceleration gain, and subscripts \(o\) and \(n\) denote the old and new followers, respectively.
}

\section{MOMDP Formulation}
\label{Background}
This section formulates the multi-objective Markov Decision Process (MOMDPs) for the considered problem. We then discuss the design of state-action space and rewards function.
The state transitions and rewards are a function of the  AV  environment and actions taken by the AV.

{
\subsection{MOMDPs and Pareto Front}
\label{subsection:morl_momdps}
The goal of MORL is to obtain policies among $M$ conflicting or competing objectives, where the relative importance (preferences) of each objective may be known or unknown to the agent. Similar to RL, MORL can be formulated by MOMDP which extends the MDP by defining a new reward space, preference space, and preference function, i.e., $\mathcal{M} = <\mathcal{S}, \mathcal{A},  \mathbf{r}, \mathcal{P}, \Omega, \iota_0> $, where $ \mathbf{r} \in \mathbb{R}^{H}$ is a vector of reward functions corresponding to $H$ objectives, e.g., $\mathbf{r} = [ r^1,r^2,\dots,r^H]$, $ \Omega$ is the preference space where $ \boldsymbol{\omega} \in \Omega$ is the preference vector corresponding to $H$ objectives, and $\sum_{h=1}^{H}{{\omega}}^h = 1$.  $\iota_0$ is the probability distribution over initial states.
In MOMDPs, a policy $\pi: \mathcal{S} \to \mathcal{A}$ defines a mapping from states to actions with the goal of maximizing a vector of expected rewards.
$\mathcal{P}(s_{t+1} | s_t,a_t)$ indicates the transition probability for the agent to take an action $a_t \in \mathcal{A} $ on state $s_t \in \mathcal{S} $ to the next state  $s_{t+1}\in \mathcal{S} $ in time step $t$.
Given the distribution $\iota_0$ and a policy $\pi$, the expected discounted return is defined as:
  \begin{equation}
  \label{eq:bellman-update-envelope}
    \mathbf{Q}_{\pi}(s,a, \boldsymbol{\omega})  = \mathbb{E}_{\pi} \left[  \mathbf{r}(s_t,a_t)  +  \mathcal{\gamma}  \mathbf{Q}_{\pi}(s_{t-1},a_{t-1}, \boldsymbol{\omega}) \right]
   \end{equation}  
where  
$\mathbf{r}(s_t,a_t)$ is the immediate vector valued reward at time step $t$ for $H$ objectives. Maximizing the expected reward involves solving the following MOO problem
$\max \mathbf{Q}_\pi = \max_{\pi} [ Q_{\pi}^1,Q_{\pi}^2,\dots,Q_{\pi}^H]$.

\begin{figure}
\begin{subfigure}{0.5\textwidth}
\includegraphics[width=\linewidth]{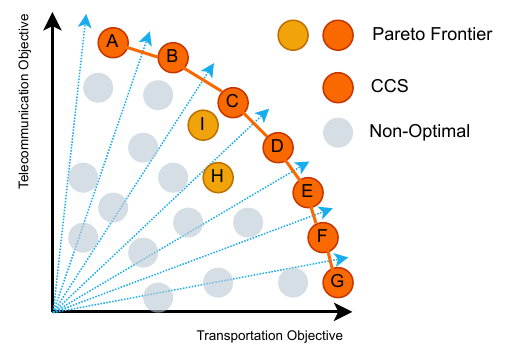} 
\caption*{(a)}
\end{subfigure}
\hfill 
\begin{subfigure}{0.5\textwidth}
\includegraphics[width=\linewidth]{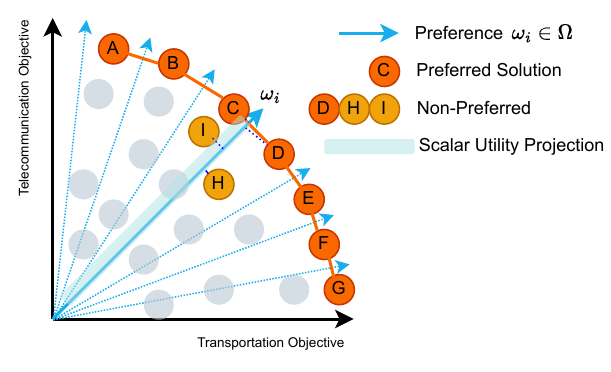} 
\caption*{(b)}
\end{subfigure}
\caption{Pareto Frontier and CCS Analysis}
\label{fig:ccs}
\end{figure}

A policy $\pi$ strictly dominates another policy $\pi'$ if $\pi$ achieves values at least as high as $\pi'$ in all objectives and strictly higher in at least one objective:
\begin{equation}
\label{eq:ppf}
\pi > \pi' \iff \forall h, {V}^h_{\pi} \geq {V}^h_{\pi'} \land \exists \: h, {V}^h_{\pi} > {V}^h_{\pi'}
\end{equation}
Furthermore, a policy $\pi$ weakly dominates another policy $\pi'$, if $\pi$ achieves values greater than or equal to $\pi'$ in all objectives, i.e., $\pi \geq \pi'$, when $ {V}^h_{\pi}  \geq {V}^h_{\pi'}, \quad  \forall h$. A policy $\pi$ is considered Pareto-optimal (or non-dominated) if it is not strictly dominated by any other policies. 

Considering all returns from MOMDP, we have Pareto frontier set $\mathcal{F}^* := \left\{\mathbf{\hat r} \mid  \nexists \: \mathbf{\hat r}' \geq \mathbf{\hat r} \right\}$ \cite{yang2019generalized}, where $\mathbf{\hat r} = \sum_{t=0}^{\infty} \gamma \cdot \mathbf{r}(s_t,a_t)$. 
For all possible preferences in $\Omega$, we define a convex coverage set (CCS) of the Pareto frontier which contains all returns that provide the maximum cumulative reward, i.e., 
\begin{equation}
\label{eq:CCS}
    \mathrm{CCS} := \left\{\mathbf{\hat r} \in \mathcal{F}^* \mid \exists	{\boldsymbol{\omega}} \in \Omega , \forall \mathbf{ \hat r}' \in \mathcal{F}^* \text{ s.t.} \: \boldsymbol{\omega}^{T}\mathbf{\hat r} \geq \boldsymbol{\omega}^{T} \mathbf{\hat r}'\right\}
\end{equation}
where $(\cdot)^{T}$ denotes the transpose operator.

{

From Fig. \ref{fig:ccs}(a), the Pareto frontier includes points A to H, encapsulating some local concave regions. Grey points indicate non-optimal solutions that do not belong to the CCS in terms of transportation and telecommunication objective.  In Fig. \ref{fig:ccs}(b), linear preferences are used to select the optimal solution from the CCS based on the highest utility. This utility corresponds to the projection length of each point onto the preference vector $\omega_i \in \Omega$. Blue arrows represent different linear preferences reflecting trade-offs between transportation and telecommunication objectives. Among the possible returns, point C achieves a higher cumulative utility compared to points D, I, and H when projected along the preference vector represented by the solid line $\omega_i$. $\omega_i \cdot \hat{r}_C >\omega_i \cdot \hat{r}_D>\omega_i \cdot \hat{r}_I>\omega_i \cdot \hat{r}_H$. the optimal solution is the point in the CCS with the largest projection along the preference vector $\omega_i$
}
}

\subsection{State Space}

The state space consists of kinematics-related features, which is a $M_1 \times F$ array that describes $F \to \{ x_j, y_j, v_j, \psi_j, n_R^j, n_T^j  \}$ specific features of AVs. We consider $M_1$ target AVs and $M_2$ surrounding AVs.  Each target AV is characterized by its (1) coordinates $(x_j,y_j)$, (2) forward velocity $v_j$, (3) heading $\psi_j$, and (4) $n_R^j $ and $n_T^j$ which are the number of RBSs and TBSs that makes AV achieves the desired data rate  in a predefined radius from its current position, respectively. 
Accordingly, the aggregated state space $\mathcal{S}$  at any time step $t$ is given by:
\begin{equation*}
\mathcal{S} =
\begin{bmatrix}
x_1 & y_1 & v_1 & \psi_1 & n_R^1 & n_T^1 \\
\vdots & \vdots & \vdots & \vdots & \vdots & \vdots \\
x_{M_1} & y_{M_1} & v_{M_1} & \psi_{M_1} & n_R^{M_1} & n_T^{M_1}
\end{bmatrix}
\end{equation*}

\subsection{Two Dimensional Action Space}
\label{subsection:2daction}
The action space consists of self-driving action space $\mathcal{A}_\mathrm{tran}$ and telecommunication action space $\mathcal{A}_\mathrm{tele}$, 
{which include 5 and 3 discrete actions, respectively.} For each time step $t$, the AV must select 
driving-related action and telecommunication-related action from action space, as shown below:
\begin{equation*}
\mathcal{A} =
\begin{bmatrix}
\{ a_{\rm tele}^1 , a_{\rm tran}^1\}  & \{ a_{\rm tele}^1 , a_{\rm tran}^2 \}  & \dots & \{ a_{\rm tele}^1 , a_{\rm tran}^5 \} \\
\vdots & \vdots & \vdots & \vdots  \\
\{ a_{\rm tele}^3 , a_{\rm tran}^1\}  & \{ a_{\rm tele}^3 , a_{\rm tran}^2 \}  & \dots & \{ a_{\rm tele}^3 , a_{\rm tran}^5 \} 
\end{bmatrix}
\end{equation*}

Note that $\mathcal{A}_{\rm tran}=\{a_{\rm tran}^1,\ldots,a_{\rm tran}^5\}$, where $a_{\rm tran}^1,a_{\rm tran}^2$ and $a_{\rm tran}^3$ indicate that AV changes its lane to the left, maintains the same lane, and changes its lane to the right, respectively. $a_{\rm tran}^4$ and $a_{\rm tran}^5$ indicate the acceleration and deceleration of AV within the same lane. It is important to note that the acceleration and deceleration rates are dynamically determined by the model in Section \ref{transportation_model}. With that being said, each AV selects the same actions does not imply that they will perform identical accelerations/deceleration. The communication action space is represented as $\mathcal{A}_{\rm tele}=\{ a_{\rm tele}^1,a_{\rm tele}^2,a_{\rm tele}^3 \}$. $a_{\rm tele}^1$ indicates scenarios where 
AV selects a BS by maximizing \textit{weighted data rate metric} (defined by equation (\ref{eq:weight_data_rate}) in Section \ref{tele_model}), which encourages traffic load balancing between BSs and discourages unnecessary HOs, especially for TBSs. In $a_{\rm tele}^2,$ the AV selects a BS with maximum $\text{WR}_{{ij}}$ by substituting $\mu=0$, if $Q_i \geq n_i$. Otherwise, AV recursively selects the next vacant best-performing BS in terms of $\text{WR}_{ij}$. 
 In $a_{\rm tele}^3$, the AV  chooses to connect to a BS with the maximum  data rate $R_{ij}$.

\subsection{Rewards}
\label{subsection:rewards}
The design of the associated reward is directly related to optimizing the driving policy and network selection, and is critical for accelerating the
convergence of the model. Generally, the AV is given a positive reward when it receives a higher HO-aware data rate while guaranteeing
safe driving. By taking any other actions that may lead to an increase in HOs, collisions, or traffic violations, the AV receives a penalty. We define the transportation reward as \cite{highway-env}:
{
\begin{equation}
\label{eq:tran_reward}
    r^{\mathrm{j,tran}}_{t}=  c_1 \left( \frac{v^j_t -v_\mathrm{min}}{v_\mathrm{max}-v_\mathrm{min}} \right)- c_2 \cdot \delta_2 + c_3 \cdot \delta_3 + c_4 \cdot \delta_4  , 
\end{equation}
}
where $v_t^j,v_{\min}$ and $v_{\max}$ are the current longitudinal velocity for AV $j$ on time $t$, the minimum and maximum speed limits, 
and {$\delta_2, \delta_3, \delta_4 $ is the collision indicator, AV right lane indicator, on road indicator, respectively}. $c_1$ and $c_2$ are the weights that adjust the value of the AV transportation reward with its collision penalty. {$c_1$ indicates that the reward received when driving at full speed, linearly mapped to zero for lower speeds.}
{
$c_3$ shows that AV was rewarded for driving on the right-most lanes, and linearly mapped to zero for other lanes. $c_4$ is the on-road reward factor, which penalize the AV for driving off highway. 
}
It is important to note that negative rewards are not allowed since they might encourage the agent to prioritize ending an episode early by causing a collision instead of taking the risk of receiving a negative return if no satisfactory trajectory is available.

For the telecommunication side, we define the reward for AV $j$ associated with BS $i^*$ at time step $t$ as:
\begin{equation}
\label{eq:tele_reward}
    r^{\mathrm{j,tele}}_t= c_5 \text{WR}_{i^*,j,t} \left(1- \text{min}(1,\xi^j_t )\right),
\end{equation}
where  $c_5$ is the coefficient to scalarlize weighted datarate. $\text{WR}_{i^*,j,t} $ is the achievable data rate compute by (\ref{eq:weight_data_rate})  and  $\xi^j_t$ is the HO probability of AV $j$ computed by dividing the number of HOs accounted until the current time $t$ by the time duration of previous time slots in the episode\footnote{Note that $c_1 \dots c_5$ are the  weights to set the priority of each term. For instance, $c_2$ needs to be sufficiently large compared to {other coefficients} for collision avoidance. {The highest penalty applies to vehicle collision.}}.

Based on the instantaneous reward, we compute the accumulated rewards, which is the summation of discounted reward among all target AVs on the highway in each training episode.
The expected return is defined as follows: 
\begin{equation}
    \mathbf{Q}_\pi (s,a, \boldsymbol{\omega}) = \mathbb{E}_{\pi} \left[\sum_{j=1}^{M_1} r^{\mathrm{j,tran}}_{t},\sum_{j=1}^{M_1} r^{\mathrm{j,tele}}_{t} \right]
\end{equation}
Our MOMDP optimal strategy for maximizing the expected reward involves the simultaneous maximization of both transportation and telecommunications objectives, i.e., $    \max_{\pi}{\mathbf{Q}_\pi (s,a, \boldsymbol{\omega})}$.

\section{Multi-Policy Envelope MORL Algorithm}
\label{morl}
In contrast to conventional DRL,  MORL requires the agent to optimize multiple objectives simultaneously. These objectives might have predefined preferences, or the preferences could be unknown.  A multiple-policy envelope solution for MORL is proposed in Section \ref{morl} for unknown preferences. The single policy solutions to the MORL problem with predefined preferences are discussed in { \textbf{Appendix}}. Although single-policy methods are adequate when we possess prior knowledge of task preferences, the acquired policy is constrained in its adaptability to situations with varying preferences. For instance, collision avoidance may not remain a high priority in highway environments with reduced vehicle density. Also, in traffic jams or parking lots where AVs are still, the preference for telecommunication rewards becomes higher. 

{  Our approach separates the model training and evaluation phases. The MORL agent is { trained offline} using a predefined simulated environment. During the evaluation stage,   we use a \emph{pool} of pretrained MORL models, each tailored to a predefined configuration defined by a given number of AVs (representing traffic conditions),  number of RF/THz BSs, and a specific topology.  This approach is also adopted by 3GPP \cite{3gpp_ai_ml_model_transfer}, i.e., at run time, the controller simply selects the  MORL model that best matches the observed snapshot, achieving zero online training overhead \cite{3gpp_ai_ml_model_transfer}. }


\begin{figure*}[t]
\centering
\includegraphics[width=\linewidth]{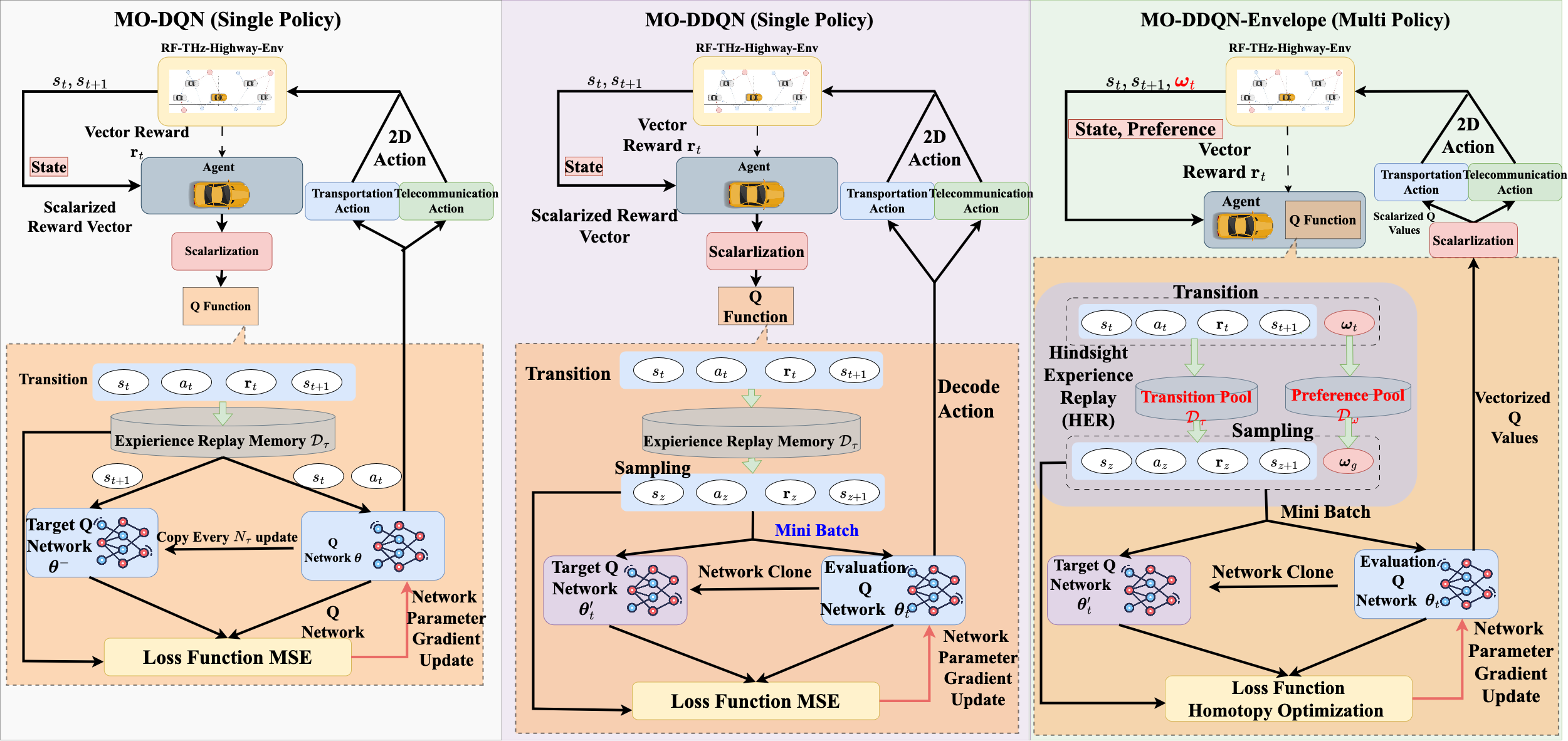} 
\vspace{-2mm}
\caption {Comparison of MO-DQN,  MO-DDQN, and the proposed MO-DDQN-envelope framework }
\label{fig:framework_comparison}
\end{figure*}

In contrast to single-policy methods, multi-policy MORL methods optimize different objectives simultaneously by maximizing a vector of rewards associated with these objectives. Our proposed MORL framework reduces reliance on predefined preferences and scalar reward combinations, enabling dynamic adjustment to associated tasks featuring distinct preferences. This approach is effective in identifying Pareto-optimal policies when preferences are unknown.

{ Our proposed MO-DDQN-Envelope algorithm is designed to learn a spectrum of Pareto-optimal policies simultaneously within a preference space ${\Omega}$, as described in Section \ref{subsection:morl_momdps} and illustrated in Fig.~\ref{fig:framework_comparison}. Unlike the Envelope-MOQ model proposed in \cite{yang2019generalized}, which employs the REINFORCE algorithm, our MO-DDQN-Envelope algorithm incorporates DDQN to enhance both convergence stability and sample training efficiency. REINFORCE, as a policy gradient method, often suffers from high variance in complex environments like \textit{RF-Thz-Highway-Env}, leading to unstable updates. In contrast, DDQN leverages temporal difference (TD) learning, which reduces variance and mitigates the overestimation bias.
Furthermore, DDQN employs replay experience to enable the agent to learn from past experiences multiple times, improving sample efficiency, and uses target networks to stabilize learning by ensuring consistent targets during updates. 
}

{During each time step, observation information is captured in the \textit{RF-THz-Highway} environment. From this observation, the tuple $\{s_t, s_{t+1}, \boldsymbol{\omega}_t\}$ is computed.}
 Following states information acquisition, the hindsight experience replay (HER) technique is employed to sample preference weights from the replay preference pool $\mathcal{D}_{\mathcal{T}}$. Then, homotopy optimization is applied to execute gradient descent, as indicated in (\ref{eq:homotopy_loss_function}). Subsequently, we perform  $Q$ network clone from evaluation network to target network periodically for every $N^-$ steps.  Notably, unlike prior single policy MORL approaches that scalarize rewards before the experience replay, MO-DDQN-envelope scalarizes rewards {after} gradient descent.
We elaborate on the Bellman operator update phase, the HER phase and homotopy optimization phase in detail in what follows.

\subsubsection{Bellman Operation with Optimal Filter}

 In the context described in Section \ref{subsection:morl_momdps} and referenced by \cite{yang2019generalized}, the expected discounted return under a policy $\pi$ is defined as $\mathbf{Q}_{\pi}(s,a, \boldsymbol{\omega}) = \mathbb{E}_{\pi} \left[ \mathbf{r}(s_t,a_t) + \gamma \mathbf{Q}_{\pi}(s_{t-1},a_{t-1}, \boldsymbol{\omega}) \right]$. 
 Yang $\textit{et al.}$ in \cite{yang2019generalized} further introduces the concept of an optimal filter $\mathcal{H}$\footnote{The optimal filter $\mathcal{H}$ is instrumental in solving the convex envelope of PPF, which represents the current solution frontier. This process is key in optimizing the Q-function, $\mathbf{Q}_{\pi}$ for a given state $s$ and preference weights $\boldsymbol{\omega}$.}, which is applied to $\mathbf{Q}_{\pi}(s,a, \boldsymbol{\omega})$ to obtain $(\mathcal{H}\mathbf{Q})_{\pi}(s,a, \boldsymbol{\omega}) = \arg_Q \sup_{a \in \mathcal{A},\boldsymbol{\omega}' \in \Omega} \mathbf{Q}_{\pi}(s,a, \boldsymbol{\omega}')$. 
 The $\arg_Q$ represents a multi-objective supremum value, ensuring that $(a,\boldsymbol{\omega}')$ achieves the maximum supremum across actions in space $\mathcal{A}$ and states $\boldsymbol{\omega}'$ within the space $\Omega$. 
Consequently, we  utilize (\ref{eq:bellman-update-envelope}) to focus the optimization on actions solely dependent on $\mathcal{H}$. The MO optimality operator can thus be defined as:
\begin{equation}
    \mathbf{Q}(s,a, \boldsymbol{\omega})  = \mathbb{E}_{s_{t+1}} \left[  \mathbf{r}(s_t,a_t)  +  \mathcal{\gamma}  (\mathcal{H}\mathbf{Q})(s_{t+1}, \boldsymbol{\omega}) \right]
    \label{eq:mo-optimal-operator}
\end{equation} 

{ 
During the training phase, $\boldsymbol{\omega}$ values are collected through free exploration of the environment. As can be seen in Fig.~\ref{fig:ccs}, $\omega_i$ specifies preference weights to balance objectives.
In the initial free exploration phase, $\omega_i$ uniformly samples from the sampling space $\Omega$.
Uniform sampling ensures that the exploration covers a broad range of trade-offs, increasing the likelihood of finding Pareto-optimal solutions across the entire objective space.

Unlike scalarized single-policy approaches discussed in \textbf{Appendix}, which fail to adapt the scalar utility across varying $\boldsymbol{\omega}$, the convex envelope formulation explicitly leverages the supremum operator to optimize across all possible actions $a$ and preference vectors $\boldsymbol{\omega}'$. 
}

\subsubsection{Hindsight Experience Replay (HER)}
HER is a method to train a RL agent to achieve multiple preferences to serve multiple objectives \cite{yang2019generalized, andrychowicz2017hindsight}. The RL agent follows a policy based on a randomly selected goal in each episode and uses the previous trajectory to update other goals simultaneously. 

In our enhanced MO-DDQN-envelope network, leveraging HER, we employ the sampling process from two distinct replay pools $\mathcal{D}_\mathcal{T}$ and $\mathcal{D}_\omega$, targeting both transition mini-batches and preference vectors.
{
We extract $N_\mathcal{T}$ mini-batch transitions,$ (s_z,a_z,\mathbf{r}_z,s_{z+1})$ to form replay buffer pool $\mathcal{D}_\mathcal{T}$, such as $ (s_z,a_z,\mathbf{r}_z,s_{z+1}) \sim \mathcal{D}_\mathcal{T}$, where $z \in [1,N_\mathcal{T}]$. 
Concurrently, we sample preference vectors $\omega_g$ in $\mathcal{D}_\omega$ to form replay buffer $ \mathcal{W} \equiv \{ \omega_g \sim \mathcal{D}_\omega \}$, with $g \in [1,N_\omega]$, $N_\omega$ indicates the count of preference weights in $\mathcal{W}$. }
Therefore, the agent AV can replay the trajectories with any preferences using "hindsight" since preferences only impact agent AVs actions rather than highway environment dynamics\cite{yang2019generalized}.

\subsubsection{Homotopy Optimization}
Our goal is to generate a single model which adapts the entire pareto frontier space $\Omega$. 
By sampling $N_\mathcal{T}$ transitions $(s_z, a_z, \mathbf{r}_z, s_{z+1} )$ and $N_\mathcal{\omega}$ preference weights $\mathcal{W} =\{  \omega_g \sim \mathcal{D}_\omega\}$ in respective replay buffer $\mathcal{D}_\tau$ and $\mathcal{D}_\omega$, we define MO-DDQN-envelope element-wise target function \cite{yang2019generalized} as follows:
\begin{equation}
\label{eq:mo_envelope_ddqn_target}
    \hat{\mathbf{Q}}(s_z, a_z, \mathbf{r}_z, s_{z+1}, \omega_g) =   \mathbf{r}_z + 
\gamma \max_{a' \in \mathcal{A}, \boldsymbol{\omega}' \in \mathcal{W}}[{\omega}_g]^{T}\mathbf{Q}(s_{z+1},a',\boldsymbol{\omega}')
\end{equation}
{for $\forall z \in [1,N_\mathcal{T}]$ and $\forall g \in [1,N_\omega]$. Finding the optimal preference weight $\boldsymbol{\omega}'$ in $\Omega$ can be an NP-hard problem due to the size and complexity of $\Omega$. Instead, finding the optimal preference $\boldsymbol{\omega}'$ in $\mathcal{W}$ is feasible.} 
{
In contrast to $\Omega$, $\mathcal{W}$ contains only those preferences that align with optimal solutions. 
This is because $\mathcal{W}$ corresponds to the convex boundary of the Pareto frontier, where the utility projection $\boldsymbol{\omega}^\top \mathbf{Q}(s,a,\boldsymbol{\omega}')$ achieves its maximum for some preference $\boldsymbol{\omega}' \in \Omega$.
Consequently, $\mathcal{W}$ serves as a reduced yet comprehensive subset of $\Omega$, facilitating efficient exploration and optimization without excluding any optimal solutions.
From a computational perspective, the use of $\mathcal{W}$ simplifies the optimization process. By maintaining the envelope $\sup_{\boldsymbol{\omega}'} \boldsymbol{\omega}^\top \mathbf{Q}(s,a,\boldsymbol{\omega}')$, the algorithm effectively focuses on preferences that yield maximal utility, avoiding unnecessary evaluations in suboptimal regions of $\Omega$. This ensures that the updated policies and preferences are always aligned with the optimal frontier. 
 Furthermore, the iterative refinement of $\mathcal{W}$ through envelope updates enables the model to integrate information from previously explored trajectories stored in the preference pool $D_\omega$. This process results in faster convergence and improved sample efficiency compared to directly exploring the entire space $\Omega$. By focusing optimization within $\mathcal{W}$, the approach avoids unnecessary evaluations in suboptimal regions of $\Omega$, thereby improving computational efficiency. 
}

By replay sampling transition $(s_t,a_t,\mathbf{r}_t,s_{t+1})$ across $N_\mathcal{T}$ transitions, we acquire the empirical estimate target function over new state $s_{t+1}$ as: 
\begin{equation}
\begin{multlined}
   \hat{\mathbf{Q}}(s_t,a_t,\boldsymbol{\omega}_t;\boldsymbol{\theta}_t') = 
   \mathbb{E}_{s_{t+1}} [ \mathbf{r}_t + \gamma \arg_Q \max_{a_t,\boldsymbol{\omega}_t'} \boldsymbol{\omega}^{T} \\
   \mathbf{Q}(s_{t+1},a_t,\boldsymbol{\omega}_t'; \boldsymbol{\theta}_t')]
\end{multlined}
\end{equation}
where $\mathbf{Q}(\cdot)$ revisits (\ref{eq:mo-optimal-operator}). 
To ensure the correctness of the training for the target value $\hat{\mathbf{Q}}$, which should be as close as possible to the actual value ($\mathbf{Q}$).
The loss function $\mathcal{L}^A(\boldsymbol{\theta}_t)$ in each time step $t$  is defined as:
\begin{equation}
\label{eq:LA_theta}
    \mathcal{L}^A(\boldsymbol{\theta}_t) =  \mathbb{E}_{s_t,a_t,\omega_t} \Bigr[|| \hat{\mathbf{Q}}(s_t,a_t,\omega_t;\boldsymbol{\theta}_t') - \mathbf{Q}(s_t,a_t,\omega_t; \boldsymbol{\theta}_t) ||^2_2 \Bigr]
\end{equation}
Since $\mathcal{L}^A(\boldsymbol{\theta}_t)$ contains many local maxima and minima, it is difficult to find the mean square error (MSE) and hard to optimize $Q_{\theta}$. To smooth the landscape of loss function $\mathcal{L}^A(\boldsymbol{\theta}_t)$,  we introduce the auxiliary loss function $\mathcal{L}^B(\boldsymbol{\theta}_t)$ as: 
\begin{equation}
\label{eq:LB_theta}
    \mathcal{L}^B(\boldsymbol{\theta}_t) =  \mathbb{E}_{s_t,a_t,\omega_t} \Bigr[|\omega_t^{T}  \hat{\mathbf{Q}}(s_t,a_t,\omega_t;\boldsymbol{\theta}_t') - \omega_t^{T } \mathbf{Q}(s_t,a_t,\omega_t; \boldsymbol{\theta}_t)| \Bigr]
\end{equation}
$\mathcal{L}^B(\boldsymbol{\theta}_t)$ contributes smooth policy adaptation for enhancing training efficiency with fewer spikes. $\mathcal{L}^B(\boldsymbol{\theta}_t)$ is advantageous for boosting agent training, but not as good for accurate approximation as $\mathcal{L}^A(\boldsymbol{\theta}_t)$ \cite{yang2019generalized}. 
Both $\mathcal{L}^A(\boldsymbol{\theta}_t)$ and $\mathcal{L}^B(\boldsymbol{\theta}_t)$  are averaged over $\omega_t$ which highlights the sampling preference feature in the proposed algorithm. However, specific weight $\omega_t$ in the past training is not directly applied to the target state-action transitions. The proposed MO-DDQN-envelope can reevaluate past transitions in $\mathcal{D}_\mathcal{T}$ with later new preferences to enhance learning efficiency and sample utilization.

Combining   (\ref{eq:LA_theta}) and (\ref{eq:LB_theta}),  we generate loss function 
\begin{equation}
\label{eq:homotopy_loss_function}
    \mathcal{L}(\boldsymbol{\theta}_t) = (1- \lambda_t) \mathcal{L}^A(\boldsymbol{\theta}_t) + \lambda_t \mathcal{L}^B(\boldsymbol{\theta}_t)
\end{equation}

\begin{figure}
    \centering
    \includegraphics[width=\linewidth]{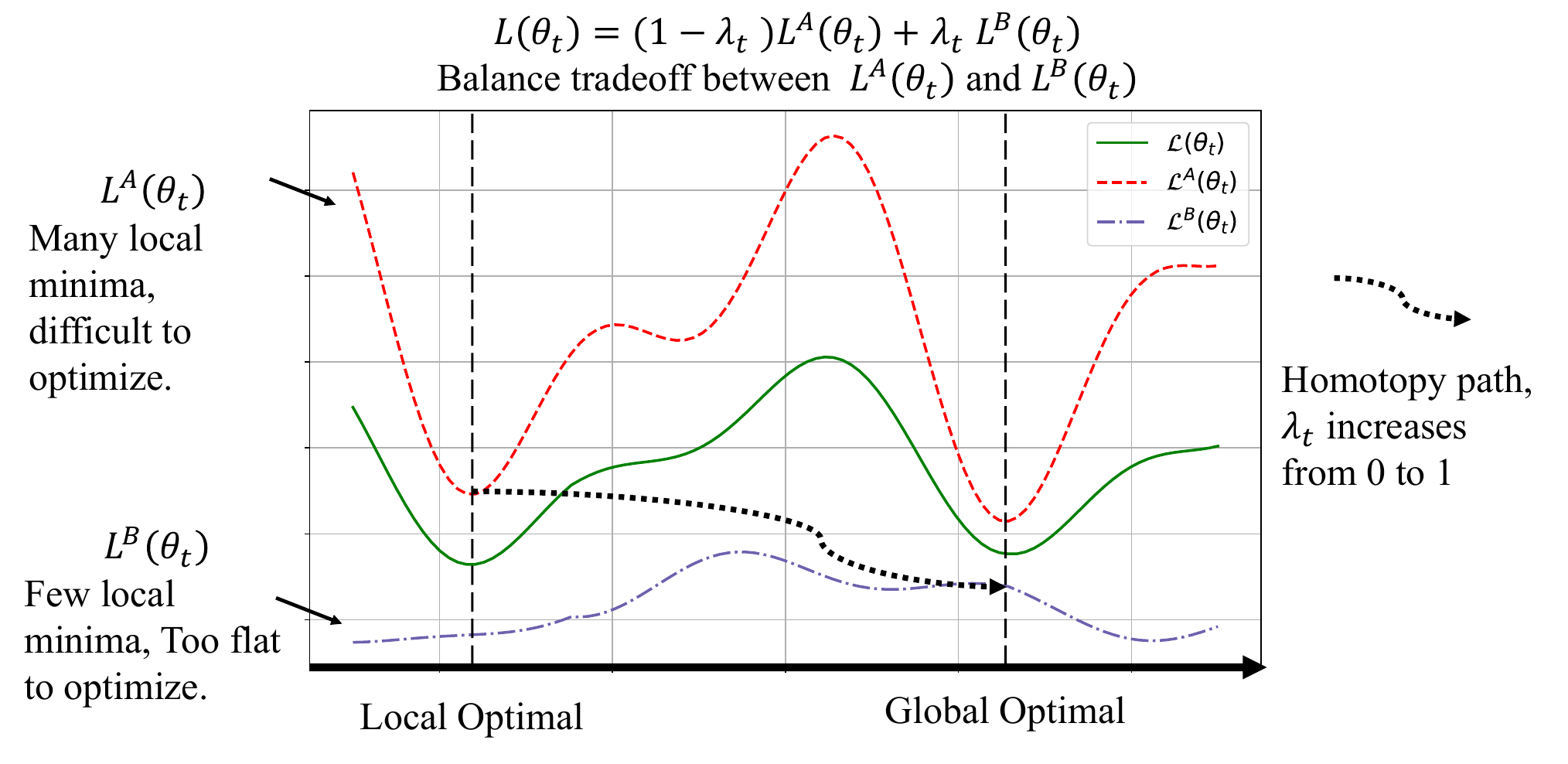} 
    \caption{An explanation for homotopy optimization method used in the MO-DDQN-Envelope}
    \label{fig:homotopy}
\end{figure}

{

 We progressively increase the value of $\lambda_t$ from $0$ to $1$ throughout the training process, thereby transitioning the loss function from $\mathcal{L}^A(\boldsymbol{\theta}_t)$ to $\mathcal{L}^B(\boldsymbol{\theta}_t)$.
During training, $\lambda_t$ continuously evolves within the range $0 < \lambda_t < 1$, ensuring a gradual shift that balances the contributions of the two objectives as the training advances \cite{watson1989modern,yang2019generalized}. 
This method called {\em homotopy optimization}~\cite{watson1989modern} is effective since for each update step, it uses the optimization result from the previous step as the initial guess. In the envelope deep MORL algorithm, $\mathcal{L}^A(\boldsymbol{\theta}_t)$ first ensures the prediction of $\mathrm{Q}$ is close to any real expected total reward, though not necessarily optimal. And, then $\mathcal{L}^B(\boldsymbol{\theta}_t)$ can  pull the current guess along the direction with better utility. 
 As depicted in Figure \ref{fig:homotopy}, the MSE loss $\mathcal{L}^A(\boldsymbol{\theta}_t)$ is difficult to optimize since there are many local minima over its landscape. Although the loss of the value metric $\mathcal{L}^B(\boldsymbol{\theta}_t)$ has fewer local minima, it is also difficult for optimization since there are many vectors $\mathrm{Q}$  minimizing value metric $\omega_t^{T} \cdot| \hat{\mathbf{Q}}(s_t,a_t,\omega_t;\boldsymbol{\theta}_t') - \mathbf{Q}(s_t,a_t,\omega_t; \boldsymbol{\theta}_t)|$, making $\mathcal{L}^B(\boldsymbol{\theta}_t)$ is flat. The homotopy path connecting $\mathcal{L}^A(\boldsymbol{\theta}_t)$ and $\mathcal{L}^B(\boldsymbol{\theta}_t)$ provide better opportunities to find the optimal global parameters $\mathcal{L}(\boldsymbol{\theta}_t)$ }.

We first trying to reduce the discrepancy between target and estimate $\mathbf{Q}$ value as $(\hat{\mathbf{Q}}(s_t,a_t,\omega_t;\boldsymbol{\theta}_t) - \mathbf{Q}(s_t,a_t,\omega_t; \boldsymbol{\theta}_t)$ and then taking gradient  descent $\nabla_{\boldsymbol{\theta}_t}$ for estimate $\mathbf{Q}$ value to adjust direction to reduce MSE. 
Consequently, parameter for MO-DDQN-envelope will be updated as,
\begin{equation}
\label{eq:gradient_descent_mo_envelope_ddqn}
\begin{multlined}
\boldsymbol{\theta}_{t+1}  =  \boldsymbol{\theta}_{t} +  \mathbb{E}_{s_t,a_t,s_{t+1}} [ (\hat{\mathbf{Q}}(s_t,a_t,\omega_t;\boldsymbol{\theta}_t) \\
-  \mathbf{Q}(s_t,a_t,\omega_t; \boldsymbol{\theta}_t) )^{T} \nabla_{\boldsymbol{\theta}_t} \mathbf{Q}(s_t,a_t,\omega_t; \boldsymbol{\theta}_t)  ] 
\end{multlined}
\end{equation}
We reset our target $Q$-network with evaluation $Q$-network every $N^-$ steps, i.e., $\theta \gets  \theta'$.
The training algorithm of the proposed MO-DDQN-envelope is shown in \textbf{Algorithm} \ref{algo2}. 
\begin{algorithm}[t]
\caption{Multi-Objective Envelope DDQN}
\label{algo2}
\SetAlgoLined
\KwResult{Learned action-value function $\mathbf{Q}_\theta$ and Policy $\pi$}
\KwData{Evaluation $Q$-network $\mathbf{Q}_\theta$, Target $Q$-network $\mathbf{Q}_{\theta'}$, Preference sampling pool $\mathcal{D}_\omega$, HER transition sampling pool $\mathcal{D}_\mathcal{T}$, Balance weight path $p_\lambda$}

\textbf{Initialization:}\\
HER replay buffer $\mathcal{D}_\mathcal{T} \gets \emptyset$,\\
Initialize Q-network weights $\theta$ randomly,\\
Initialize target Q-network weights $\theta' \gets \theta$,\\
Initialize $Q(s,a)$ for all states $s$ and actions $a$, including AVs, TBSs, RBSs.

\While{episode $<$ episode limit and runtime $<$ time limit}{
  Initialize $t \gets 0$ and state $s_t$ based on environment\\
\While{$t \leq T_{hl}$ }{ 
    \For{ Target AV $j$ \textbf{from} 1 \textbf{to} $M_1$}{
    $a_t$ select action from $\mathcal{A}$ with probability of $\epsilon$ or Select $a_t$ from $a_t = \arg\max_{a \in \mathcal{A}}{\boldsymbol{\omega}^T} \mathbf{Q}(s_t,a,\boldsymbol{\omega};\boldsymbol{\theta}_t)$ with probability of $ 1-\epsilon$. \\
    Derive $a^{\mathrm{tran}}_{t}$ and $a^{\mathrm{tele}}_{t}$ from $a_t$;\\
    Apply $a^{\mathrm{tran}}_{t}$ and $a^{\mathrm{tele}}_{t}$ to target AV $j$;\\ Observe vector reward $\mathbf{r}_t$ and next state $s_{t+1}$;
    }
    \If{update neural network}{
      Store $(s_t, a_t, \mathbf{r}_t, s_{t+1})$ in $\mathcal{D}_\mathcal{T}$;\\
\colorbox{cyan!20}{\parbox{\dimexpr\linewidth-17\fboxsep\relax}{
      \textbf{Hindsight Experience Replay (HER):} \\
      $ \{ (s_z, a_z, \mathbf{r}_z, s_{z+1})  \sim \mathcal{D}_\mathcal{T} \}$;\\
Sample $N_\omega$ preferences $ \mathcal{W} =\{  \omega_g \sim \mathcal{D}_\omega\}$; 

      \textbf{Bellman Update:} \\
        Compute $\hat{\mathbf{Q}}(s_z, a_z, \mathbf{r}_z, s_{z+1}, \omega_g)$ for each sampled  transition and preference:
      \[
      \begin{cases}
      \mathbf{r}_z, & \text{if } s_{z+1} \text{ is terminal}\\
       (\ref{eq:mo_envelope_ddqn_target}), & \text{otherwise} 
      \end{cases}
      \]
      $ \forall z \in [1,N_\mathcal{T}]$ and $\forall g \in [1,N_\omega]$
      }}
\colorbox{green!20}{\parbox{\dimexpr\linewidth-17\fboxsep\relax}{
      \textbf{Homotopy Optimization:}\\
      Update $\mathbf{Q}_\theta$ by minimizing the loss with \\ gradient descent by (\ref{eq:homotopy_loss_function});\\
      Gradually increase $\lambda$ following the path $p_\lambda$;}}
    }
Update target $\mathbf{Q}_{\theta'}$ weights $\theta' \gets \theta$ by (\ref{eq:gradient_descent_mo_envelope_ddqn}) every $N^-$ steps;
  }
     $t \gets t + 1$\\
  Compute policy $\pi$ based on learned $\mathbf{Q}_\theta$;
}
\end{algorithm}

\subsection{Complexity Analysis}
\subsubsection{Training Complexity Analysis}
{ 
We first analyze the main loop of \textbf{Algorithm} \ref{algo2}. The key components contributing to the time complexity include:

\begin{itemize}
    \item \textit{Action Selection:} We consider the horizon limit for each episode, denoted as $T_{\text{hl}}$. The process of selecting an action at each timestep incurs a time complexity of $\mathcal{O}(|\mathcal{A}|)$ for each target AV, where $|\mathcal{A}|$ represents the size of the action space. Thus, the time complexity is $\mathcal{O}( M_1 \cdot T_{\text{hl}} \cdot |\mathcal{A}|)$ .
    \item \textit{Hindsight Experience Replay (HER):} Sampling $N_\mathcal{T}$ transitions from the HER replay buffer and $N_\omega$ preferences results in a time complexity of $\mathcal{O}(N_\mathcal{T} \cdot N_\omega)$.
    \item \textit{Homotopy Optimization:} This phase involves a Fully Connected Neural Network (FCNN) consisting of an input layer, an output layer, and $E$ fully connected hidden layers. Let $N_e$ denote the number of neurons in the $e$-th fully connected layer. The time complexity for this phase is $\mathcal{O}\left(\sum_{e=1}^{E+1}(N_{e-1}\cdot N_{e})\right)$, accounting for the computational cost for each layer's connections.
    \item \textit{Policy Adaptation:} Utilizing a DDQN, the input layer's neurons correspond to the dimensionality of the state space $\mathcal{S}$, and the output layer's neurons correspond to the action space $\mathcal{A}$. Thus, the numbers of neurons in the input and output layers are $N_0 = |\mathcal{S}|$ and $N_{E+1} = |\mathcal{A}|$, respectively. The time complexity for this phase is $\mathcal{O}(|\mathcal{S}|  \cdot |\mathcal{A}|)$, which can be considered constant upon convergence.
\end{itemize}

{ Therefore, the overall training complexity for MO-DDQN-envelope on 1 episode can be expressed as $\mathcal{O}(M_1 \cdot T_{\text{hl}} \cdot |\mathcal{A}| + N_\mathcal{T} \cdot N_\omega + \sum_{e=1}^{E+1}(N_{e-1}\cdot N_{e}) + |\mathcal{S}| \cdot |\mathcal{A}|)$. }
}
\subsubsection{Evaluation Complexity Analysis}
{ 
Assume we have \( N_l \) layers in the neural network, each containing \( N_u \) neurons. According to Section III-B, the neural network processes \( |\mathcal{S}| \) inputs from the state space and produces \( |\mathcal{A}| \) possible actions as output. The overall complexity for a forward pass through the network is given by:
\[
\mathcal{O}(|\mathcal{S}| + |\mathcal{A}| + N_u^{N_l}) \approx \mathcal{O}(N_u^{N_l}),
\]
and \( N_u^{N_l} \), the computation for the hidden layers, dominates the complexity. For evaluation, we assume a single step is implemented with a constant-time complexity of \( \mathcal{O}(1) \). If there are \( M_1 \) target AVs and each travels for at most \( T_{hl} \) time steps in one episode, the total number of steps in an episode is \( M_1 \cdot T_{hl} \). Thus, the overall complexity for one simulation episode is:
\[
\mathcal{O}(M_1 \cdot T_{hl} \cdot (N_u^{N_l} + \mathcal{O}(1))) \approx \mathcal{O}(M_1 \cdot T_{hl} \cdot N_u^{N_l}),
\]
where the constant \(\mathcal{O}(1)\) is negligible compared to the forward pass complexity.

}

\section{Simulation and Performance Evaluation}
In this section, we demonstrate the performance of the proposed algorithms and highlight the complex dynamics between wireless connectivity, traffic flow, and AV’s speed. All Experiments are executed on a PC equipped with Windows 11, Intel i7-8770 CPU 3.2 GHz and 16 GB DDR5, AMD RX580 8GB GDDR5. Additionally, \textit{Google Colab}, is the cloud platform employed for reproduction and verification.
\label{simu}
\subsection{Simulation Environment}

Our proposed simulation framework is composed of three main components: 
\begin{itemize} 
    \item \textbf{Telecommunication and Transportation Environment:} 
    We introduce the \textit{MO-RF-THz-Highway-env} framework\footnote{https://github.com/sunnyyzj/highway-env-1.7}, an enhanced version of \textit{highway-env}~\cite{highway-env}, designed to support both autonomous driving policy and 5G/6G network selection for multiple AVs.
    \item \textbf{Extended MO-Gymnasium:} We extend \textit{MO-Gymnasium}\footnote{https://github.com/sunnyyzj/MO-Gymnasium} \cite{Alegre+2022bnaic} to provide an application programming interface (API) to communicate between DRL algorithms and  \textit{MO-RF-THz-Highway-env}.
    \item \textbf{MORL Algorithms Simulation:}
    For single policy MORL, simulation utilizes modified \textit{rl-agents}\footnote{https://github.com/sunnyyzj/rl-agents} from \cite{rl-agents}. For multi-policy MORL,  
    the proposed MO-DDQN-envelope\footnote{https://github.com/sunnyyzj/morl-baselines} and the other multi-policy algorithms simulation are extended from \textit{MORL-baselines} in \cite{morl_baselines}.
\end{itemize}
The \textit{MO-RF-THz-Highway-env} features five one-way lanes, each with a length of 1500m and a width of 4m, as depicted in Figure~\ref{fig:BS_distributions}. In our experimental setup, a default configuration consists of 5 target AVs and 20 surrounding AVs, each having a length of 5m. The longitudinal velocity of each AV ranges from 10 m/s to 35 m/s. At the beginning of each episode, these AVs are randomly placed across the five lanes. Along both sides of the highway, 5 RBSs and 10 to 50 TBSs are also randomly positioned to ensure a non-uniform distribution at the beginning of each episode. This random placement strategy aims to facilitate the examination of MORL training effectiveness across various VNets and traffic scenarios, as opposed to a singular, common scenario. The maximum duration for each episode is set to 30 time steps. An episode is considered \textit{collision-free} if it meets two criteria: absence of collisions among all target AVs, and maintenance of a high weighted data rate regarding (\ref{eq:weight_data_rate}) during travel through the episode. 
For single policy MORL, we set rewards coefficients { in (\ref{eq:tran_reward}) (\ref{eq:tele_reward}) from Section \ref{subsection:rewards}}, $c_1,c_2,c_3,c_4,c_5$ { set} to $\num{0.4},1,0.1,0.2,\num{4.5e-7}$, respectively. 
{
The collision coefficient $c_2$ is set significantly higher than the others because collision avoidance is our highest-priority and incurs the greatest penalty \cite{dong2023comprehensive,leurent2020safe}. On the other hand, the achievable handover-aware data rate ranges from \num{5e7} bits per second (bit/s) to \num{5e8} bit/s. To balance the transportation and telecommunication objectives effectively, $c_5$ is set to $4.5 \times 10^{-7}$, ensuring that the data rate is appropriately scaled for optimizing weighted-sum single policy MORL.
}

{We utilize a Multi-Layer Perceptron (MLP) to construct the $Q_\theta$ neural network for the training and evaluation phases. The architecture consists of three fully connected layers (FNNs), each containing 128 neurons. Following these layers, we apply the ReLU activation function to improve training efficiency.
The output layer of the target policy network $Q_\theta'$ employs a sigmoid activation function to constrain the output range of actions. This ensures that the action values remain within a feasible and interpretable range, which is critical for MOO decision-making tasks. This architecture was chosen after a comprehensive hyperparameter tuning process.}
{ This configuration strikes a balance between model accuracy, training efficiency, and convergence performance.}
Simulation parameters are detailed in Table~\ref{tab:simulation_parameter} unless otherwise specified.

\begin{table}
	\centering

	\begin{tabular}{m{6.2cm}<{\raggedleft} m{1.8cm}<{\raggedright}}  
		\hline 
		\textbf{Parameter} & \textbf{Value} \\  
		\hline
		\multicolumn{2}{c}{ {Value used in system model}} \\
		\hline

        RBSs frequency $(f_R)$ &  3.5 GHz \\
        TBSs frequency $(f_T)$ &  1 THz \\
        {Maximum number of affordable AVs quota for a single RBS, TBS $(Q_R),(Q_T) $} & 5, 10 \\
        Antenna gain for TBSs and RBSs $(G_{T}^{\mathrm{tx}})$, $(G_{T}^{\mathrm{rx}})$ & 316.2 \\
        RBSs channel bandwidth $(W_R)$ &  $\num{4e7}$ \\
        TBSs channel bandwidth $(W_T)$ &  $\num{5e8}$ \\
	Transmit powers of RBSs and TBSs   $(P_{R}^{\mathrm{tx}})$, $(P_{T}^{\mathrm{tx}})$ & 1~W \\
        Molecular absorption coefficient $(K_a)$ & 0.05~m$^{-1}$ \\
        Path loss exponent $(\alpha)$ & 4 \\
        Length for each AV $(l_j)$ &  5~m \\
        {Target AV heading and lateral control gain $(K_j^{\psi})$,$(K_j^{y})$} &  5, $\frac{5}{3}$\\
        Maximum AV steering angle for AV $j$ $(\max{\beta_j})$ &  $\frac{\pi}{3}$ \\
        Surrounded AV $j$ desired maximum acceleration and deceleration $(a_j)$ &  3 m/s , -5 m/s \\
        Acceleration reduction factor $(\delta_a)$ & 4 \\
        Number of lanes $(N_L)$ & 5 \\
		\hline
		\multicolumn{2}{c}{{Value used in MORL}}\\
		\hline
  	Learning rate $(\alpha_{l})$ &  $\num{3e-4}$ \\
        Discount factor $(\gamma)$ & 0.995\\
        Size of the hidden layers of the value NN & $[256, 256, 256, 256]$ \\
		Epsilon decay parameter $(\epsilon)$ & 0.1 \\
        MO-DDQN-envelope HER transition pool size $(N_\mathcal{T})$ & $\num{2e6}$ \\
        The number of weight vectors to sample for the envelope target $(N_\omega)$ & 4 \\
        Frequency for cloning evaluation to target network $(N^-)$ & 200 \\
        Episode horizon limit $T_{hl}$ & 30 \\

		\hline  
	\end{tabular}
    \caption{Values of system parameters in experiments}
	\label{tab:simulation_parameter}
\end{table}

{
We evaluated all methods across five test instances, varying key variables: desired minimum and maximum longitudinal velocities ($v_{\text{min}}, v_{\text{max}}$), the number of TBSs ($n_T$), and the number of AVs ($M$). These combinations are given in \tableref{tab:test_instances}.

\begin{table}\centering 
    \vspace{-0.1cm}
    \renewcommand{\arraystretch}{1.2}
    \setlength\tabcolsep{0.8em}
    \begin{tabular}{cccccc}
    \toprule
    \textbf{Instance} & $\boldsymbol{v_{\text{min}}}$ & $\boldsymbol{v_{\text{max}}}$ & $\boldsymbol{n_T}$ & $\boldsymbol{M}$ \\
    \midrule
    I-(20,30,20,20) & 20 m/s & 30 m/s & 20 & 20 \\
    I-(25,35,20,20) & 25 m/s & 35 m/s & 20 & 20 \\
    I-(20,30,10,20) & 20 m/s & 30 m/s & 10 & 20 \\
    I-(20,30,20,50) & 20 m/s & 30 m/s & 20 & 50 \\
    I-(30,40,20,20) & 30 m/s & 40 m/s & 20 & 20 \\
    \bottomrule
    \end{tabular}
    \caption{Test Instances with Parameters.}
    \label{tab:test_instances}
    \vspace{-0.1cm}
\end{table}
}

 \subsection{Baselines and Evaluation Metrics}

{

To comprehensively evaluate the performance of MO-DDQN-envelope, we implement the followsing baseline algorithms for comparison: MO-DQN, MO-DDQN (as discussed in \textbf{Appendix}), MO-dueling-DDQN, and MO-PPO. MO-dueling-DDQN and MO-PPO are variations of MO-DDQN, utilizing single-policy multi-objective dueling DDQN and Proximal Policy Optimization (PPO) algorithms, respectively, instead of DDQN. These algorithms, specifically developed for performance evaluation, are extensions of single-policy RL methods originally proposed in \cite{wang2016dueling} and \cite{schulman2017proximal}, respectively. Additionally, we incorporate several state-of-the-art algorithms from MORL-Baselines \cite{morl_baselines} for comparison. These include Scalarized Q-learning for single-policy MORL (MO-Q) \cite{van2013scalarized}, Scalarized DQN for single-policy DQN (MO-DQN) \cite{10001396}, Action Branching Architectures with Dueling DDQN for MORL (MO-Dueling-DDQN) \cite{yan2023multi}, Multi-objective exploration for proximal policy optimization (MO-PPO) \cite{khoi2021multi}, and Expected Utility Policy Gradient (EUPG) \cite{roijers2018multi}.
}
To evaluate the performance, we consider the following  metrics. Assume episode $e$ ends on time step $T_e$. we define \textbf{(1)}  total transportation reward: $R^{tran}_e = \mathbb{E}_{j \in M_1}[ \sum_{t=1}^{T_e}r^{\mathrm{j,tran}}_{t}] $. \textbf{(2)} total telecommunication reward: $R^{tele}_e =  \mathbb{E}_{j \in M_1}[ \sum_{t=1}^{T_e}r^{\mathrm{j,tele}}_{t}] $. \textbf{(3)} collision rate: $\delta_e =  1-\frac{T_e}{T_{hl}} $ \textbf{(4)} HOs Probability: $\xi_e =  \mathbb{E}_{j \in M_1} [\xi^j_{T_e}]$. {where $r^{\mathrm{j,tran}}_{t}$ and $ r^{\mathrm{j,tele}}_{t}$ are instantaneous transportation and telecommunication rewards specified in equations (\ref{eq:tran_reward}) and (\ref{eq:tele_reward}), respectively. $T_{hl}$ represents the horizon limit of each episode. $\xi^j_{T_e}$ denotes  HOs probability by the end of each episode ($T_e$'s step) defined in equation (\ref{eq:tele_reward}). }
{
Table~\ref{tab:evaluation_models} evaluate total transportation reward \eqref{eq:tran_reward}, total telecommunication reward \eqref{eq:tele_reward}, total rewards defined by $R^{total}_e = R^{tele}_e +R^{tran}_e$ over MO-DQN, MO-DDQN, and MO-DDQN-Envelope.
}
\begin{table*}\centering \footnotesize
    \vspace{-0.1cm}

    \setlength\tabcolsep{0.8em}
\resizebox{0.9\linewidth}{!}{
    \begin{tabular}{cccccc}
    \toprule
    \toprule
     \multicolumn{1}{c}{\shortstack{Target Velocity}} & \multicolumn{1}{c}{\multirow{2}{*}{Model}} & \multicolumn{4}{c}{Number of AVs $M$} \\
     \cmidrule(lr){3-6}
     \multicolumn{1}{c}{(m/s)} & \multicolumn{1}{c}{} & \multicolumn{1}{c}{$M$=10} & \multicolumn{1}{c}{$M$=20} & \multicolumn{1}{c}{$M$=30} & \multicolumn{1}{c}{$M$=40} \\
    \midrule
    \multirow{3}{*}{15} & MO-DQN & 23.42/10.21/33.63 & 22.71/9.18/31.89 & 21.09/7.45/28.54 & 18.35/6.12/24.47 \\
    & MO-DDQN & 24.28/12.31/36.59 & \textbf{24.89/11.41/36.30} & 21.85/8.63/30.48 & 19.41/7.11/26.52 \\
    & MO-DDQN-Envelope & \textbf{25.76/13.45/39.21} & 23.74/10.03/33.77 & \textbf{23.16/9.86/33.02} & \textbf{20.83/7.89/28.72} \\
    \midrule
    \multirow{3}{*}{20} & MO-DQN & 20.85/8.56/29.41 & 19.92/6.74/26.66 & 19.23/6.35/25.58 & 17.48/4.39/21.87 \\
    & MO-DDQN & 21.62/10.19/31.81 & 21.04/8.57/29.61 & 20.11/7.93/28.04 & 18.72/5.97/24.69 \\
    & MO-DDQN-Envelope & \textbf{22.49/11.45/33.94} & \textbf{21.98/9.92/31.90} & \textbf{20.87/8.72/29.59} & \textbf{19.61/6.18/25.79} \\
    \midrule
    \multirow{3}{*}{25} & MO-DQN & 18.21/7.48/25.69 & 17.45/5.91/23.36 & 16.34/5.12/21.46 & 14.87/4.37/19.24 \\
    & MO-DDQN & 18.92/8.19/27.11 & 17.88/6.72/24.60 & 16.97/5.94/22.91 & 15.32/4.92/20.24 \\
    & MO-DDQN-Envelope & \textbf{19.64/9.01/28.65} & \textbf{18.37/7.51/25.88} & \textbf{17.04/6.52/23.56} & \textbf{16.01/5.38/21.39} \\
    \midrule
    \multirow{3}{*}{30} & MO-DQN & 16.09/6.32/22.41 & 15.76/5.03/20.79 & 14.92/4.76/19.68 & 13.81/4.21/18.02 \\
    & MO-DDQN & 16.72/6.89/23.61 & 16.21/5.32/21.53 & 15.43/5.14/20.57 & \textbf{14.89/4.91/19.80} \\
    & MO-DDQN-Envelope & \textbf{17.35/7.45/24.80} & \textbf{16.88/5.97/22.85} & \textbf{15.97/5.63/21.60} & 14.12/4.49/18.61 \\
    \bottomrule
    \bottomrule
    \end{tabular}
    }
    \caption{ \footnotesize Evaluation performance of MO-DQN, MO-DDQN, and MO-DDQN-Envelope (total transportation reward / total telecommunication reward / total reward).}
        \label{tab:evaluation_models}
    \vspace{-0.1cm}
\end{table*}

\begin{figure*}
\includegraphics[width=\linewidth]{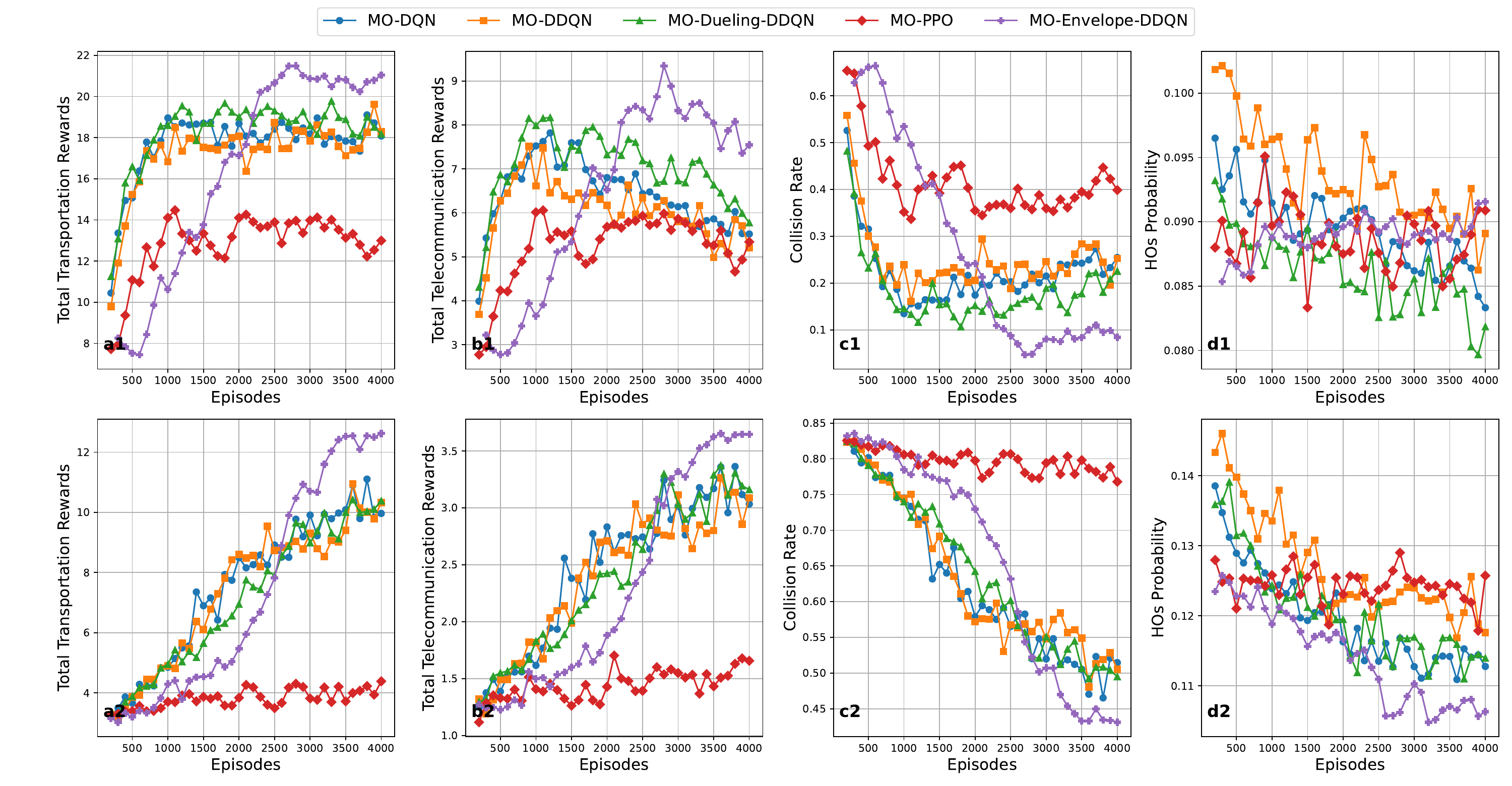} 
\centering

\caption{\footnotesize Training performance on (a) total transportation rewards, (b) total telecommunication rewards, (c) collision rate, and (d) HOs probability. $n_T=20$, the desired minimum and maximum longitudinal velocities are $v_{\text{min}}=20$m/s and $v_{\text{max}}=30$m/s, respectively, for the top row and $v_{\text{min}}=30$m/s and $v_{\text{max}}=40$m/s, respectively, for the bottom row, and number of AVs are 20 and 50 in the top and bottom row, respectively}

\label{fig:envelope_ddqn_dqn_training}
\end{figure*}

\begin{figure*}
\includegraphics[width=\linewidth]{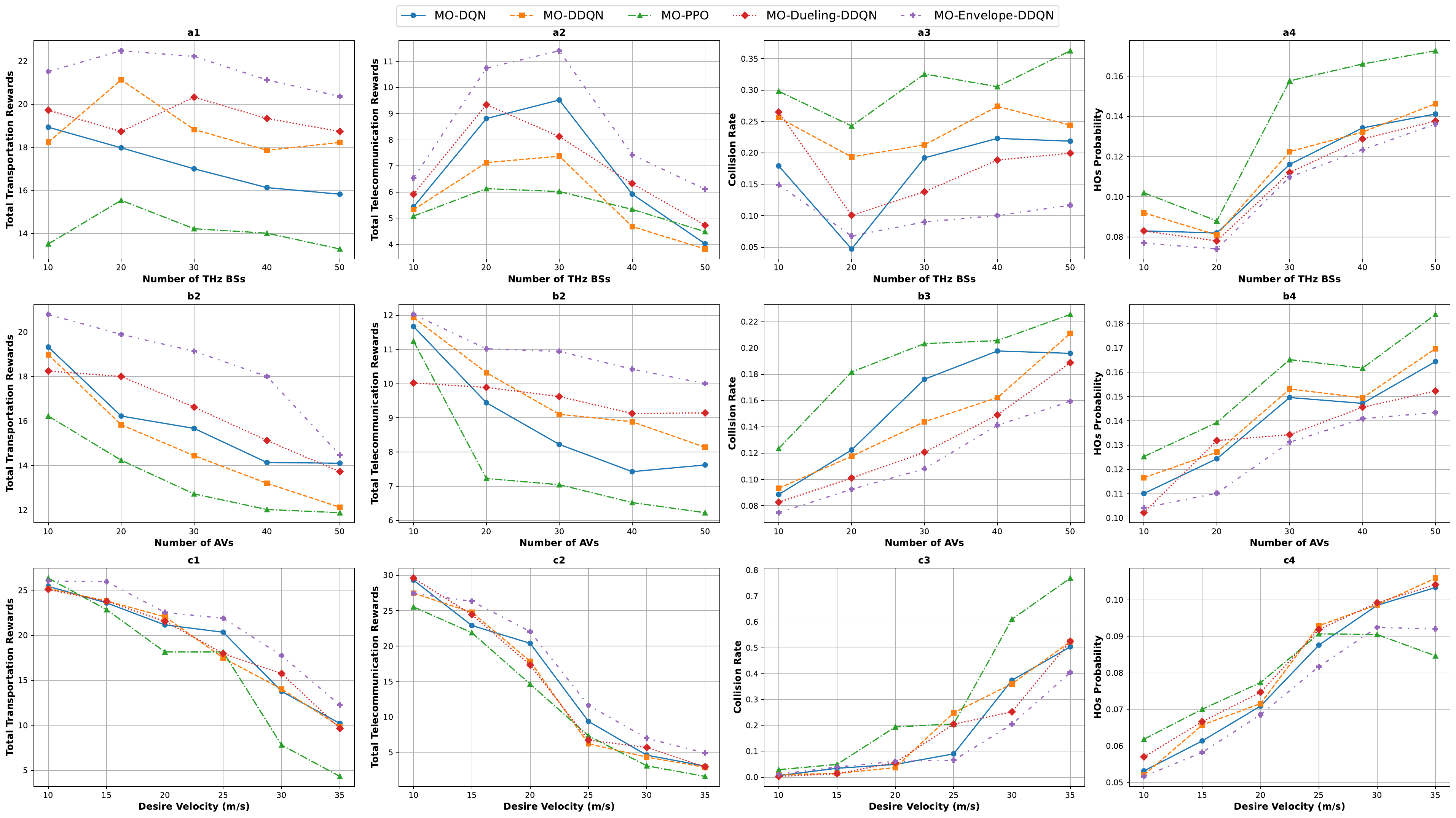} 
\centering

\caption{\footnotesize Evaluation performance on (1) total transportation rewards, (2) total telecommunication rewards, (3) collision rate, and (4) HOs probability, as a function of  (a) variation in TBSs with counts ranging from 10 to 50 while maintaining an AV velocity of 25 m/s for 20 AVs, and (b) variation in vehicle numbers adjusting from 10 to 50 with a fixed AV velocity of 25 m/s alongside 20 TBSs. {(c) variation in desired velocity from 5 to 30 m/s with fixed 20 AVs and 20 TBSs.}}

\label{fig:envelope_ddqn_dqn_evaluation_varies_tbss}
\end{figure*}

\subsection{Results and Discussions}
 \subsubsection{Training Performance}
We examine the training performance of the proposed MO-DDQN-envelope algorithm and compare it with other baselines of MORL approaches. 
Fig.~\ref{fig:envelope_ddqn_dqn_training} {depicts total transportation rewards, total telecommunication reward, collision Rate, and HOs probability} as a function of desired velocity of AV. MO-DDQN-envelope performs better than benchmark algorithms (i.e. MO-DQN and MO-DDQN) when evaluating performance over every 100 episodes. Shown by learning curves, we note that MO-DDQN-envelope has slower convergence to higher total cumulative training rewards no matter transportation and telecommunication sides. However, MO-DDQN-envelope algorithm  balances both transportation and telecommunication objectives. Collision rate and HOs probability reduce better than the other benchmarks. 
To better understand this  improvement, recall MO-DDQN-envelope samples experience from the replay buffer which contains recent past preferences and rarely new exploration preferences. Past preferences are based on the weight vector $\Vec{\omega}$ in terms of transportation and telecommunication objectives $ \Vec{\omega} =[\omega_\mathbf{tran},\omega_\mathbf{tele}]^{T}$ which marginally improves the training for each objective individually regardless of preferences. 

For the convergence rate, thanks to \textit{Homotopy Optimization}, training first focuses on training accuracy on two objectives but later gradually focuses on faster convergence, training rewards are less viable after 3000 episodes of training than in the early stage. Also, we found the collision rate is significantly reduced from 0.7 to around 0.2, which also illustrates the training model is improving safety on the highway. 

\subsubsection{Impact of BSs Density}
\label{compare}
We evaluate the performance by averaging over 500 evaluation epochs on models after 4500 episodes of training. In each evaluation step, we randomly distribute different numbers of TBSs alongside the highway in the simulation environment. As depicted in Fig. \ref{fig:envelope_ddqn_dqn_evaluation_varies_tbss}, MO-DDQN-envelope gains an evaluation advantage compared to other benchmarks. As the number of BSs grow, the transportation rewards do not change significantly, however, the telecom rewards first increases due to better connectivity and later decreases due to more HOs.  Growing TBSs also increases the average collision rate due to potential reduction in AVs speed to maintain connectivity. 

\subsubsection{Impact of the Number of AVs}
As shown in Fig. \ref{fig:envelope_ddqn_dqn_evaluation_varies_tbss}, increasing the number of AVs leads to  more crowded highway scenarios.  Thus, more grouped AVs connect to the same BS resulting in network outages due to the maximum quota at each BS. Also, frequent lane changes and speeding on the crowded highway cause more congestion and collisions, which reduces transportation performance.

\subsubsection{Impact of Desired AV Speeds}
{Fig. \ref{fig:envelope_ddqn_dqn_evaluation_varies_tbss} depicts that slow moving AVs outperform in terms of both transportation and telecommunications. Increasing speeds lead to higher collision occurrences. Moreover, AVs at higher speeds switch BSs more frequently, incurring significant handover penalties according to (\ref{eq:tele_reward}), thus adversely affecting rewards in both domains.
}

\subsubsection{Pareto Front Analysis}
\begin{figure*}
\includegraphics[width=\linewidth]{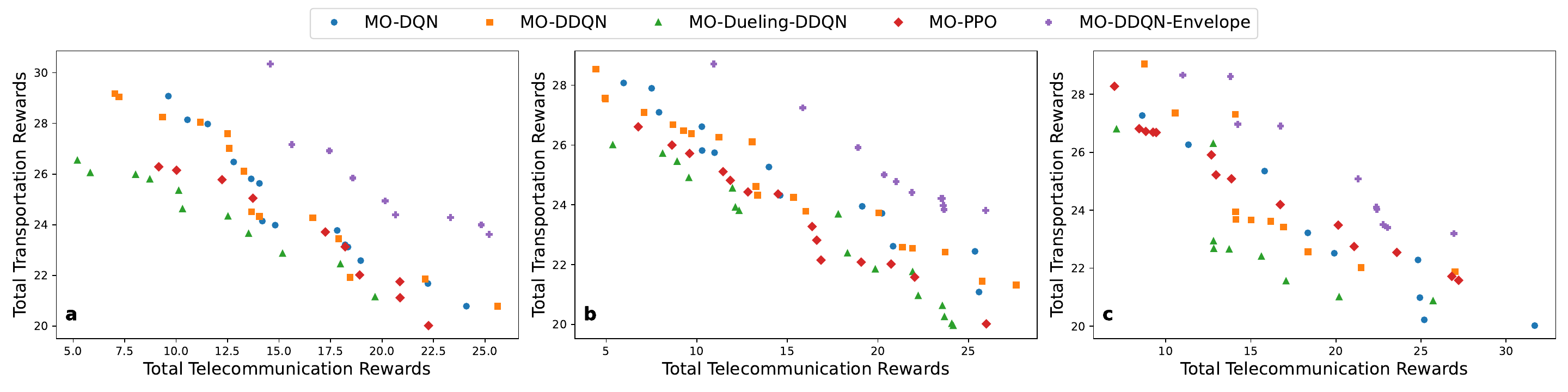} 
\centering

\caption{\footnotesize Pareto Frontier Comparison in MOO for total Transportation reward ($R^{{tran}}$) and total telecommunication reward ($R^{{tele}}$) among  MO-DQN, MO-DDQN, MO-dueling-DDQN, MO-PPO, and  MO-DDQN-Envelop, across instances:(a) I-(20,30,20,20), (b) I-(20,30,10,20), (c) I-(20,30,20,50)}

\label{fig:pareto-frontier}
\end{figure*}

We employ the CCS in  (\ref{eq:CCS}) as a means to assess the excellence of the estimated Pareto fronts. A greater Pareto frontier value indicates a closer proximity of the Pareto front to the optimal one in terms of transportation and telecommunication objectives. To compute CCS, we select performance on single policy MO-DQN and MO-DDQN as reference points.  Recall we need to maximize both transportation rewards and telecommunication rewards. The best solutions are situated in the top right corner, as depicted in Fig. \ref{fig:pareto-frontier}. Specifically, it demonstrates that our proposed algorithm MO-DDQN-envelope outperforms the MORL baselines in approximating the Pareto fronts. However, in the high-density transportation scenario, MO-DDQN yields a Pareto front of similar quality to the other baselines.

\subsubsection{Training and Evaluation Time Complexity Per Step }
{ 
We trained each algorithm for 4000 episodes, with up to $T_{hl}$ steps per episode, resulting in a total duration per step of approximately 0.643 seconds. The breakdown of time per step during training and evaluation is presented in \tableref{tab:training_evaluation}. 
The reward objectives are modified as follows:
\begin{itemize}
\item \textbf{Telecommunication Only:} This scenario considers only the telecommunication reward in the MDP design. The telecommunication reward has no impact on the training process. 

    \item \textbf{Transportation Only:} This scenario focuses solely on the transportation reward, with the telecommunication reward not influencing the training. 

    \item \textbf{{Telecommunication+Transportation:}} This scenario incorporates both the transportation and telecommunication rewards jointly during the training process.
\end{itemize}

\begin{table*}\centering \footnotesize
    \vspace{-0.1cm}
    \renewcommand{\arraystretch}{0.9}
    \setlength\tabcolsep{0.8em}
    \begin{tabular}{cccccc}
    \toprule
    \toprule
    \multicolumn{1}{c}{\textbf{Instance}} & \multicolumn{2}{c}{\textbf{Training (s)}} & \multicolumn{2}{c}{\textbf{Evaluation (s)}} \\
    \cmidrule(lr){2-3} \cmidrule(lr){4-5}
    & \textbf{Transportation} & \textbf{Telecommunication + Transportation} & \textbf{Transportation} & \textbf{Telecommunication + Transportation} \\
    \midrule
    I-(20,30,20,20) & 0.0743 & 0.0754 & 0.0643 & 0.0654 \\
    I-(25,35,20,20) & 0.0724 & 0.0731 & 0.0614 & 0.0613 \\
    I-(20,30,10,20) & 0.0764 & 0.0784 & 0.0637 & 0.0648 \\
    I-(20,30,20,50) & 0.0826 & 0.0845 & 0.0667 & 0.0672 \\
    I-(30,40,20,20) & 0.0858 & 0.0903 & 0.0725 & 0.0736 \\
    \bottomrule
    \bottomrule
    \end{tabular}
    \caption{Comparison of Training and Evaluation Times for Different Instances (in seconds). }
    \label{tab:training_evaluation}
    \vspace{-0.1cm}
\end{table*}

It is important to note that the time consumption of Telecommunication+Transportation is reasonable and comparable  to the  system policy update 0.067~s (15 Hz) \cite{highway-env}. 
In \cite{leurent2020safe,highway-env,dong2023comprehensive}, the authors considered AV motion updates at a rate of 15 Hz, i.e., {0.067}s or 67 ms.  Note that, the update frequency is not dependent on the desired velocity. On the other hand, according to uRLLC \eqref{eq:achievable_rate}, the transmission latency is given by $D_t = \frac{L_B}{W}$, where $W=W_T$ for TBSs and $W=W_R$ for RBSs, thus $D_t = 5 \times10^{-8}$s for TBSs and $D_t = 4 \times10^{-7} $s for RBSs \cite{Lancho2020Saddlepoint,Shirvanimoghaddam2019Short}, which are relatively smaller than $0.0667$s. As such, the telecommunication actions can update at the same time scale as the AV motion action updates.
Thus, updating and synchronizing telecommunication and AV motion updates in environments such as \textit{highway-env} \cite{highway-env} or the proposed \textit{RF-THz-highway-env} is reasonable.

\subsubsection{Simultaneously Multi Objective Optimization Analysis}
\begin{figure}
    \centering
    \includegraphics[width=\linewidth]{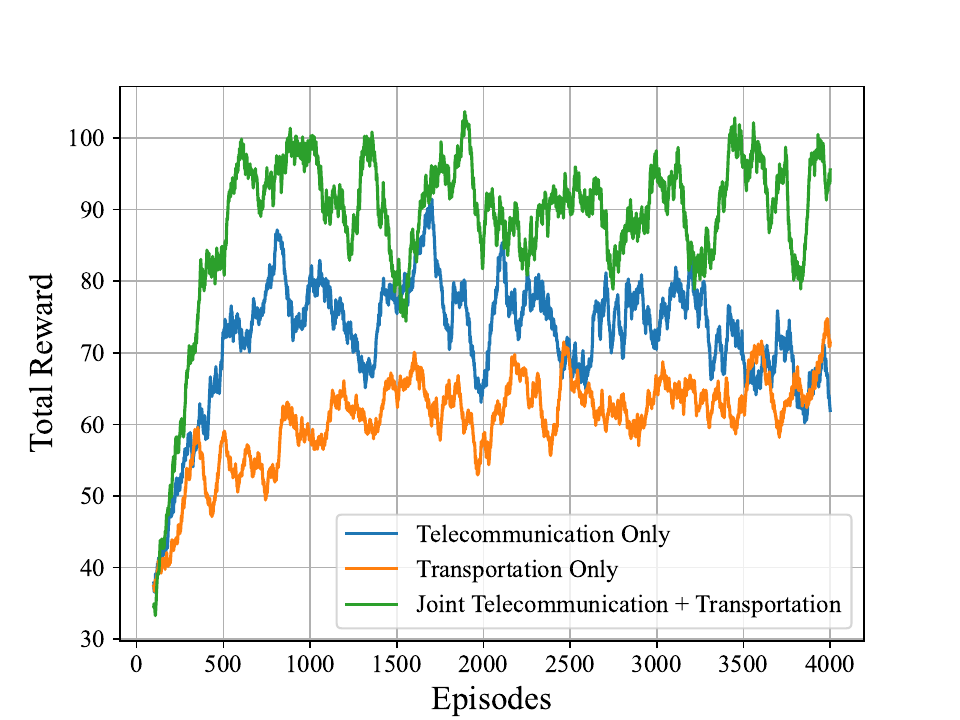} 
    \caption{ MO-Envelope-DDQN performance comparison with (1) Telecommunication  optimization only, (2) Transportation  optimization only, and (3) Joint Telecommunication + Transportation optimization.}
    \label{fig:comparison_figures}
\end{figure}
As depicted in Figure \ref{fig:comparison_figures}, we consider the  three scenarios to examine the impact of optimizing different objectives. we found that considering both telecommunication and transportation objectives performs well compared to relying on {Telecommunication Only} and {Transportation Only} alone.

}

\subsubsection{MORL Benchmarks Training Time Comparison}
\begin{table*}\centering 
    \vspace{-0.1cm}
    \renewcommand{\arraystretch}{0.9}
    \setlength\tabcolsep{0.8em}
    \begin{tabular}{ccccccc}
    \toprule
    \toprule
    \textbf{Instance} & \textbf{MO-Q\cite{van2013scalarized}} & \textbf{MO-DQN \cite{10001396}} & \textbf{MO-Dueling-DDQN \cite{yan2023multi}} & \textbf{MO-PPO \cite{khoi2021multi}} & \textbf{EUPG \cite{roijers2018multi}} & \textbf{MO-Envelope-DDQN} \\
    \midrule
    I-(20,30,20,20) & 30:36 & 31:10 & 30:45 & 31:50 & 32:05 & \textbf{30:15} \\
    I-(25,35,20,20) & 27:24 & 27:22 & 28:59 & 28:41 & 28:17 & \textbf{26:25} \\
    I-(20,30,10,20) & 30:11 & 30:53 & 30:32 & 31:22 & 31:57 & \textbf{30:06} \\
    I-(20,30,20,50) & 32:00 & 32:20 & 31:50 & 33:00 & 33:15 & \textbf{31:45} \\
    I-(30,40,20,20) & 21:45 & 22:08 & 21:33 & 22:52 & 23:35 & \textbf{21:17} \\
    \bottomrule
    \bottomrule
    \end{tabular}
    \caption{\footnotesize Comparison of Training Times for Different MORL Methods. The times reported in the tables are formatted as \textbf{MM:SS}, where \textbf{MM} represents the number of minutes and \textbf{SS} represents the number of seconds.}
    \label{tab:comparison}
    \vspace{-0.1cm}
\end{table*}

{ 

We trained each algorithm for 4000 episodes on the test instances given in \tableref{tab:test_instances} and recorded the total training time, as summarized in \tableref{tab:comparison}. Our proposed {MO-Envelope-DDQN} demonstrates competitive training times while offering significant improvements in multi-objective optimization performance. Our results in  \tableref{tab:comparison} show that the training time is mostly comparable to the baselines, if not significantly less.}

\section{Conclusion}
\label{conclude}
We introduce a novel MORL framework tailored for devising joint network selection and autonomous driving policies within a multi-band VNet. Our goals encompass enhancing traffic flow, minimizing collisions, maximizing data rates, and minimizing handoffs (HOs). We achieve this through controlling vehicle motion dynamics and network selection, employing a unique reward function that optimizes data rate, traffic flow, load balancing, and penalizes HOs and unsafe driving behaviors. The problem is formalized as a MOMDP, integrating telecommunication and autonomous driving utilities in its rewards. We propose single policy and multi-policy MORL solutions with predefined and unknown preferences.
Numerical results demonstrate the superiority of our proposed solution over weighted sum-based MORL solutions with DQN, showcasing improvements of 12.7\%, 18.9\%, and 12.3\% on average transportation reward, average telecommunication reward, and average HO rate, respectively.
{ Future research could enhance the generalization of the MORL model to better adapt to dynamic traffic conditions using well-established strategies such as meta-learning \cite{10439641}. 
}
\section*{Acknowledgment}
The authors would like to thank Dr. Hongda Wu for his valuable insights during the initial discussions and his contributions to this work's development and improvement.

\appendix
\section{Appendix}


{Given a set of preferences in MORL problems, single policy algorithms
aim to scalarize the reward value to determine the best policy, considering the relative priorities assigned to competing objectives. We explore two DRL methods: DQN  and DDQN  for MORL, 
each of which employs a 
neural network parameterized by $\boldsymbol{\theta}_t$
(in each time step $t$) to approximate the $Q$-value function for a state-action pair, i.e., $Q(s_t, a_t; \boldsymbol{\theta}_t)$. After taking action $a_t$ in state $s_t$ and receiving instant reward $r_{t+1}$, we can formulate a target $Q$ function,
\begin{equation}
  {\hat{Q}}(s_t, a_t)  = \mathbb{E}_{\pi} \left[  r_t (s_t,a_t) +  \mathcal{\gamma} Q_{\pi}(s_{t-1},a_{t-1}) \right]
\end{equation}
which is used to optimize the neural network $ \boldsymbol{\theta}_t $ using gradient descent, as given in \cite{van2016deep},
\begin{equation}
\label{eq:sorl_gradient_descent}
    \boldsymbol{\theta}_{t+1} = \boldsymbol{\theta}_t + \kappa \left( \hat{Q}(s_t, a_t) - Q(s_t, a_t; \boldsymbol{\theta}_t) \right) \nabla_{\boldsymbol{\theta}_t} Q(s_t, a_t; \boldsymbol{\theta}_t),
\end{equation}
where $\kappa$ is a positive scalar representing the learning rate. 
}
To learn a single policy for multiple tasks, we scalarize the reward vector by applying the predefined priority of each objective function \cite{10001396}, where a weighted reward function $r_t = \sum_{j=1}^{M_1} r^{\mathrm{j,tran}}_{t}+\sum_{j=1}^{M_1} r^{\mathrm{j,tele}}_{t} $
is defined to facilitate the conversion of multi-dimensional rewards into a scalar value.

{

\label{subsection:dqn}
The MO-DQN method  incorporates a target Q-network and experience replay to stabilize the learning process and ensure convergence, as discussed in the following:
\begin{itemize}
    \item \textit{Target Network:}
    Another set of neural network $\boldsymbol{\theta}_t^{-}$ is introduced to compute target $Q$ value at each time step $t$, which has the same architecture as $\boldsymbol{\theta}_t$,  but with frozen parameters. Specifically, $\boldsymbol{\theta}_t^{-}$ only copies those parameters from $\boldsymbol{\theta}_t$ every $N^-$ steps and remains fixed until the next scheduled update \cite{mnih2015human}. 
    The target value for the MO-DQN is defined as:
    \begin{equation}
    \label{eq:dqn-update}
        {\hat{Q}}
        (s_t, a_t) = r_{t+1} + \gamma \arg\max_{a} Q(s_{t+1}, a; \boldsymbol{\theta}_{t}^{-})
    \end{equation}

    \item \textit{Experience Replay:}
    To address issues related to correlations between sequential observations and to improve data efficiency, MO-DQN utilizes the experience replay mechanism, which stores past transition tuples $(s_z, a_z, s_{z+1}, r_z)$ in a replay buffer $\mathcal{D}_\mathcal{T}$ with size $N_\mathcal{T}$, {i.e., $z \in \{  1, \dots, N_\mathcal{T} \}$}. During the training phase, mini-batches of these transitions are randomly sampled from the buffer.
    This method not only reduces the variance of each update but also allows the neural network
    to benefit from learning across a diverse range of past experiences, thus avoiding local optima and overfitting.

\end{itemize}

}

{Unlike MO-DQN, where the current weights \( \boldsymbol{\theta}_t \) are used both to select and evaluate actions, MO-DDQN utilizes a separate set of parameter \( \boldsymbol{\theta}_{t}' \) to evaluate the value of the policy, ensuring a more reliable estimate by decoupling the selection and evaluation of actions. Given $\boldsymbol{\theta}_t$ and $\boldsymbol{\theta}_t'$ corresponding to evaluation and target $Q$ networks, respectively, the target value function in MO-DDQN \cite{van2016deep} is updated as follows:}
\begin{equation}
\label{eq:ddqn-update}
{\hat{Q}}(s_t, a_t) = r_{t} + \gamma Q\left(s_{t+1}, \arg\max_{a} {Q}(s_{t+1}, a; \boldsymbol{\theta}_t); \boldsymbol{\theta}_{t}'\right)
\end{equation}
{
where the action selection is guided by the online weights  $\boldsymbol{\theta}_t $. 

}

 { In MO-DQN and MO-DDQN, the neural network parameterized by $\boldsymbol{\theta}_t$ associated with evaluation function is updated by minimizing the mean square error loss $\mathcal{L}(\boldsymbol{\theta})$ between $Q$ and $\hat{Q}$ as follows \cite{mnih2015human}: 
    \begin{equation}
    \label{eq:sorl_loss}
        \mathcal{L}(\boldsymbol{\theta}_{t}) = \mathbb{E}_{z \in \mathcal{D}_\mathcal{T}} \left(Q(s_z, a_z;{\boldsymbol{\theta}_{t}}) - \hat{Q}(s_z, a_z)\right)^2
    \end{equation}
    }
 The proposed MO-DQN and MO-DDQN are illustrated in Fig. \ref{fig:framework_comparison}. The training algorithm of the proposed MO-DQN and MO-DDQN are in \textbf{Algorithm} \ref{algo1}.

\begin{algorithm}[t]
\caption{MO-DDQN Algorithm}
\label{algo1}
\SetAlgoLined
\KwResult{Learned action-value function $Q_{\boldsymbol{\theta}}$ and Policy $\pi$}
\KwData{Evaluation $Q$-network $Q$ with weights ${\boldsymbol{\theta}}$, Target $Q$-network $\hat{Q}$ with weights ${\boldsymbol{\theta}'}$  (for MO-DDQN only), Experience replay memory $\mathcal{D}_\mathcal{T}$, Mini-batch size $N_\mathcal{T}$, Horizon limit of each episode $T_{hl}$.}

\textbf{Initialization:} \\
Experience replay memory $\mathcal{D}_\mathcal{T} \gets \emptyset$, \\
Initialize $Q$-network weights $\boldsymbol{\theta}$ randomly, \\
For MO-DDQN: Initialize target network weights $\boldsymbol{\theta}' \gets \boldsymbol{\theta}$, \\
Initialize $Q(s,a)$ for all states $s$ and actions $a$, including AVs, TBSs, and RBSs.

\While{episode $<$ episode limit and runtime $<$ time limit}{
  Initialize $t \gets 0$ and state $s_t$ based on environment\\
\While{$t \leq T_{hl}$ }{ 

RL agent select $a_t$ from $\mathcal{A}$ with probability $\epsilon$ or select $a_t$ from $\max_{a \in \mathcal{A}}{Q(s_t,a_t; \boldsymbol{\theta})}$ with probability of $ 1-\epsilon$. \\
Derive $ a^{\mathrm{tran} }_{t}$ and $a^{\mathrm{tele} }_{t}$  from $a_t$\\
    Apply $a^{\text{tran}}_{t}$ and $a^{\text{tele}}_{t}$,\\ observe reward $r_t$ and next state $s_{t+1}$.\\
    Store transition $(s_t, a_t, s_{t+1}, r_t)$ in $\mathcal{D}_\mathcal{T}$.\\
    \textbf{Experience Replay:} Sample a mini-batch of transitions $(s_z, a_z, r_z, s_{z+1})$ from $\mathcal{D}_\mathcal{T}$,\\ where $z \in \{1, \ldots, N_\mathcal{T}\}$.\\
    \textbf{Set target-$Q$ for each sampled transition:}\\
    \For{each transition $z$}{
      \uIf{episode ends at step $z+1$}{
        $\hat{Q}(s_z, a_z) = r_z$
      }
      \Else{
        Use $\hat{Q}$ to compute $\hat{Q}(s_z, a_z)$ according to MO-DQN or MO-DDQN update by (\ref{eq:dqn-update}), (\ref{eq:ddqn-update}).
      }
    }
Perform a gradient descent step on (\ref{eq:sorl_loss}) with respect to network parameters $\boldsymbol{\theta}$ \\ 
    \If{MO-DDQN}{
    Update target $\hat{Q}$ weights $\boldsymbol{\theta}' \gets \boldsymbol{\theta}$  every $N^-$ steps;
    }
    $t \gets t + 1$
  }
  Update policy $\pi$ based on learned $Q$.
}
\end{algorithm}


\bibliography{main_abrv}

\begin{IEEEbiography}[{\includegraphics[width=1in,height=1.25in,clip,keepaspectratio]{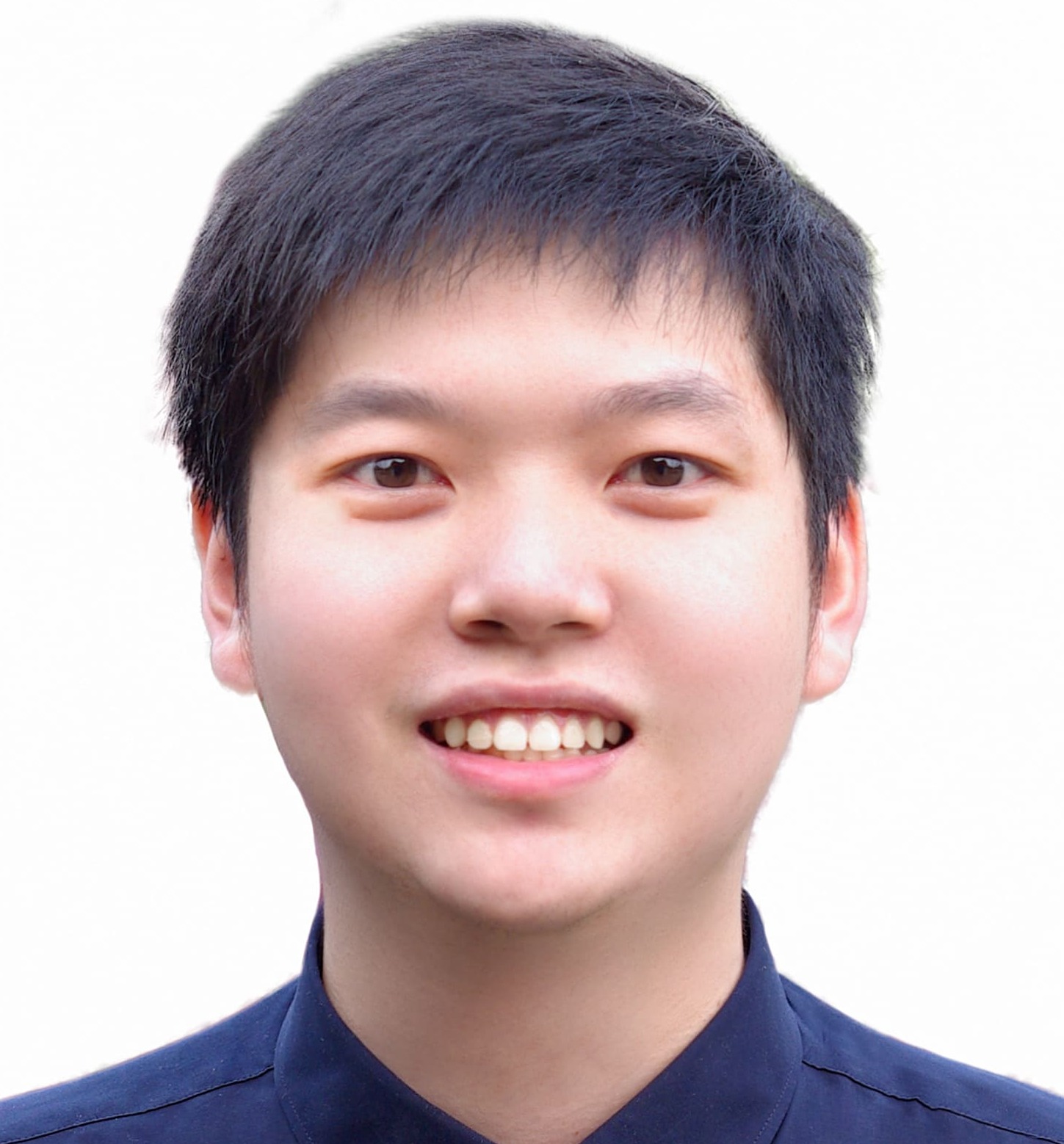}}]{Zijiang Yan} 
 (Graduate Student Member, IEEE) received the B.S. degree with a double major in Computer Science and Statistics from York University, Toronto, ON, Canada, in 2021. He is currently a Research Assistant in the Department of Electrical Engineering and Computer Science, Lassonde School of Engineering, York University. 
His research interests include AI-enabled communications, Quantum Machine Learning, Diffusion Models, and Large Language Models. He received third place in the 2025 Student Innovation Competition on Sustainable Space Communications, hosted by the Satellite and Space Communications Technical Committee (SSC TC) of the \textsc{IEEE} Communications Society (\textsc{IEEE ComSoc}). He also received the Lassonde Undergraduate Research Award (LURA) from York University in 2021. In addition, he has served as a session chair at multiple flagship conferences, such as  \textsc{IEEE GLOBECOM 2023}  and \textsc{IEEE ICC 2025}.
\end{IEEEbiography}
\begin{IEEEbiography}[{\includegraphics[width=1.25in,height=1.25in,clip,keepaspectratio]{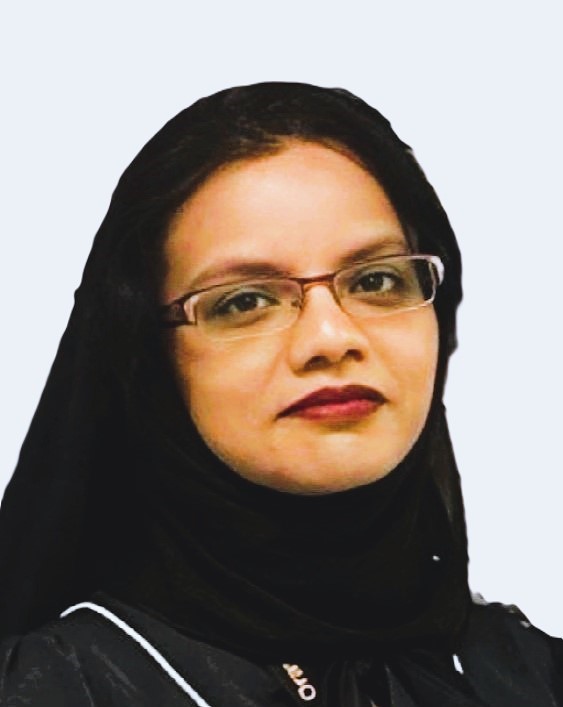}}]{Hina Tabassum} \,
(Senior Member, IEEE) \,\, (M'12--SM'18)
received the Ph.D. degree from the King Abdullah University of Science and Technology (KAUST). She is currently an Associate Professor with the Lassonde School of Engineering, York University, Canada, where she joined as an Assistant Professor in 2018. Prior to that, she was a Postdoctoral Research Associate at the University of Manitoba, Canada. She is also appointed as a Visiting Faculty at the University of Toronto in 2024 and the York Research Chair of 5G/6G-enabled mobility and sensing applications in 2023, for five years. She is also appointed as the \textsc{IEEE ComSoc} Distinguished Lecturer for the term 2025--2026. She is listed in Stanford’s list of the World’s Top Two-Percent Researchers (2021--2024). She received the Lassonde Innovation Early-Career Researcher Award in 2023 and was recognized by \textsc{N2Women} as a Rising Star in Computer Networking and Communications in 2022. She has published over 100 refereed articles in well-reputed \textsc{IEEE} journals, magazines, and conferences. 
She has received multiple Exemplary Editor awards from \textsc{IEEE Communications Letters} (2020), \textsc{IEEE Open Journal of the Communications Society (OJCOMS)} (2023--2024), and \textsc{IEEE Transactions on Green Communications and Networking} (2023). She was also named an Exemplary Reviewer (Top 2\%) for \textsc{IEEE Transactions on Communications} in 2015, 2016, 2017, 2019, and 2020.
She is the Founding Chair of the Special Interest Group on THz Communications in the \textsc{IEEE Communications Society (ComSoc)} Radio Communications Committee (RCC). She served as an Associate Editor for \textsc{IEEE Communications Letters} (2019--2023), \textsc{IEEE Open Journal of the Communications Society (OJCOMS)} (2019--2023), and \textsc{IEEE Transactions on Green Communications and Networking} (2020--2023). She is currently serving as an Area Editor for \textsc{IEEE OJCOMS}, and as an Associate Editor for \textsc{IEEE Transactions on Communications}, \textsc{IEEE Transactions on Wireless Communications}, \textsc{IEEE Transactions on Mobile Computing}, and \textsc{IEEE Communications Surveys \& Tutorials}.
\end{IEEEbiography}

\end{document}